\documentclass[journal]{IEEEtran}
\usepackage{amsmath,amsfonts,epsfig}
\usepackage{multirow,hhline,color,booktabs}
\usepackage{xcolor, soul}
\usepackage{enumitem}
\usepackage{graphicx}
\usepackage[export]{adjustbox}
\usepackage{comment}

\usepackage{url}
\usepackage{tablefootnote}
\usepackage{pifont}% http://ctan.org/pkg/pifont
\newcommand{\cmark}{\ding{51}}%
\newcommand{\xmark}{\ding{55}}%
\def\L{{\cal L}}
\newcommand{\DL} {D_{\lambda}}

\newcommand{\DS} {D_{S}}

\newcommand{\D}{\footnotesize}
\newcommand{\DD}{\tiny}

\newcommand{\bz}{\mathbf{z}}
\newcommand{\hz}{\mathbf{\hat{z}}}

\newcommand{\ru} {\rule{0mm}{3mm}}

\newcommand{\bestim}[1]{\textbf{{\color{green!60!black} #1}}}
\newcommand{\bestfiveim}[1]{{\color{green!60!black} #1}}
\newcommand{\worstfiveim}[1]{{\color{red!90!black} #1}}

\newcommand{\bestavg}[1]{{\color{green!60!black}\textbf{#1}}}
\newcommand{\bestfiveavg}[1]{{\color{green!60!black} #1}}
\newcommand{\worstfiveavg}[1]{{\color{red!90!black} #1}}

\newcommand{\git}{{\tt \url{https://github.com/matciotola/hyperspectral_pansharpening_toolbox}}}

\newcommand{\image}[1]{\includegraphics[width=0.14\linewidth]{./prisma_img/#1.jpeg}}

\newcommand{\imagerr}[1]{\includegraphics[width=0.095\linewidth]{./prisma_img/#1.jpeg}}

\begin{document}
\title{Hyperspectral Pansharpening: Critical Review, Tools and Future Perspectives}

\author{Matteo~Ciotola,
		  Giuseppe~Guarino,
          Gemine~Vivone,
          Giovanni~Poggi,
          Jocelyn~Chanussot,
          Antonio~Plaza,
          Giuseppe~Scarpa

\thanks{This work was supported by
the Italian Space Agency (ASI) under Grant ``Space It Up!'', Spoke3, CUP E63C24000220006.}        
\thanks{Matteo Ciotola, Giuseppe Guarino and Giovanni Poggi are with the Department
of Electrical Engineering and Information Technology, University Federico II, 80125 Napoli, Italy
(e-mail: giuseppe.guarino2@unina.it (G.G.), matteo.ciotola@unina.it (M.C.)
and poggi@unina.it (G.P.)).}
\thanks{Gemine Vivone is with the National Research Council, 
Institute of Methodologies for Environmental Analysis, CNR-IMAA, 85050 Tito, Italy, and also
with the National Biodiversity Future Center (NBFC), 90133 Palermo, Italy
(e-mail: gemine.vivone@imaa.cnr.it).}
\thanks{Jocelyn Chanussot is with the INRIA, CNRS, Grenoble INP, LJK, University of Grenoble Alpes, 38000 Grenoble, France (e-mail: jocelyn.chanussot@grenoble-inp.fr).}
\thanks{Antonio Plaza is with the Hyperspectral Computing Laboratory, Department of Technology of Computers and Communications, Escuela Politecnica de Caceres, 10003 Caceres, Spain (e-mail: aplaza@unex.es).}
\thanks{Giuseppe Scarpa is with the Department of Engineering, University Parthenope, 80143 Napoli, Italy 
(e-mail: giuseppe.scarpa@uniparthenope.it).}
}

% The paper headers
%\markboth{Journal of \LaTeX\ Class Files,~Vol.~13, No.~9, September~2014}%
%{Shell \MakeLowercase{\textit{et al.}}: Bare Demo of IEEEtran.cls for Journals}

\maketitle

\begin{abstract}
Hyperspectral pansharpening consists of fusing a high-resolution panchromatic band and a low-resolution hyperspectral image to obtain a new image with high resolution in both the spatial and spectral domains.
These remote sensing products are valuable for a wide range of applications, driving ever growing research efforts.
Nonetheless, results still do not meet application demands.
In part, this comes from the technical complexity of the task: compared to multispectral pansharpening, many more bands are involved, in a spectral range only partially covered by the panchromatic component and with overwhelming noise.
However, another major limiting factor is the absence of a comprehensive framework for the rapid development and accurate evaluation of new methods.
This paper attempts to address this issue.
We started by designing a dataset large and diverse enough to allow reliable training (for data-driven methods) and testing of new methods.
Then, we selected a set of state-of-the-art methods, following different approaches, characterized by promising performance, and reimplemented them in a single PyTorch framework.
Finally, we carried out a critical comparative analysis of all methods,  using the most accredited quality indicators.
The analysis highlights the main limitations of current solutions in terms of spectral/spatial quality and computational efficiency, and suggests promising research directions.
To ensure full reproducibility of the results and support future research,
the framework (comprising methods and assessment codes, and dataset setup procedures) is shared on \git, as a single Python-based reference benchmark toolbox.

\end{abstract}


\IEEEpeerreviewmaketitle

\section{Introduction}
\label{sec:intro}
High resolution is the most desired feature in a remote sensing image.
High spatial resolution enables precise detection of objects and structures.
High spectral resolution allows for the accurate classification of land covers.
Unfortunately, there is a trade-off between these desirable properties.
To obtain images with good signal-to-noise ratio, sufficient energy must be received in each acquisition cell,
and this requires bigger ground samples or large acquisition bandwidths (or both).
Modern optical systems overcome this limitation by acquiring two complementary pieces of information at the same time,
a single-band panchromatic (PAN) image with high spatial resolution and a multiband image with lower spatial resolution.
A pansharpening algorithm is then used to fuse them into a new single image
featuring the desired high resolution in both domains.

Pansharpening has been the object of intense research in recent years.
The most studied case is that of multispectral (MS) pansharpening,
which involves a limited number of bands, usually from 4 to 8, in the visible to near-infrared spectrum.
Extensive surveys on the topic can be found in the literature \cite{Vivone2014, Vivone2020, Deng2022}.
Recently, there has been a steadily growing interest in hyperspectral (HS) pansharpening,
also testified by actions such as the HS pansharpening challenge \cite{Vivone2023}.
HS images comprise hundreds of very narrow bands, covering collectively a large spectral range, going from 400 to 2500 nm.
As a result, they provide very fine-scale information on a wide variety of phenomena,
thereby holding great potential for several key applications in remote sensing,
from classification~\cite{Audebert2019,Li2019} and object detection~\cite{Zhang2020, Yan2021,Chang2022} to land use/cover mapping~\cite{Petropoulos2012,Vali2020}, crop monitoring~\cite{Fu2020,Liu2021} and estimation of land physical parameters~\cite{Niroumand2020,Pepe2020}.

%recent (2019) missions such as PRISMA ({\em PRecursore IperSpettrale della Missione Applicativa}), managed by the Italian Space Agency,
%deploys a PAN instrument coupled with the HS sensor,
%paving the road to cutting-edge research on HS pansharpening
%thanks to a data sharing policy.
From a technical point of view,
HS pansharpening is a very challenging problem\footnote{We focus on satellite-mounted systems.
Airborne missions may provide very high spatial resolution with or without pansharpening.}
because of the large resolution mismatch between PAN and HS components and the low signal-to-noise ratio involved.
The solutions designed for MS pansharpening certainly represent a good starting point also for the HS case
and their extension is often simple.
However, the notable differences between the two cases imply additional challenges that must be taken into account to prevent this naive reuse:
\begin{itemize}
\item
PAN and MS data cover approximately the same spectral range 
(visible to near-infrared) and the PAN bandwidth encloses most MS bands.
On the contrary, the spectral range of HS data goes well beyond that of the PAN (see Fig.~\ref{fig:hs-p-overlap})
and a large number of HS bands, often the majority of them, do not overlap with the PAN.
This means that, unlike for MS pansharpening, the PAN does not allow reliable prediction of the fine spatial structure of these bands.
Likewise, they should not be considered in any procedure for PAN image estimation.
\begin{figure*}
    \centering
    \includegraphics[width=0.9\linewidth]{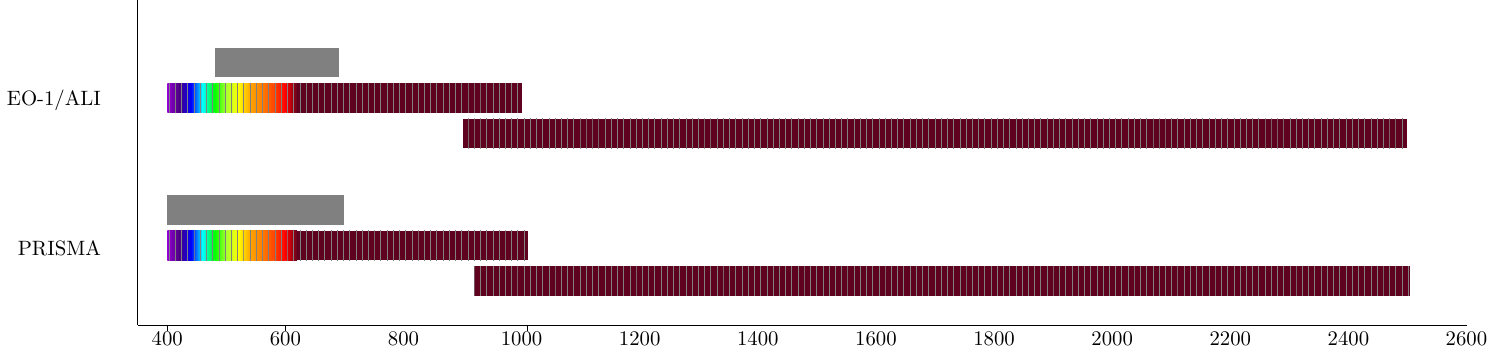}
    \caption{Spectral range of HS (colored bars) and PAN (gray bar) sensors for EO-1/ALI and PRISMA sensing systems. Both are equipped with two partially overlapped spectrometers.}
    \label{fig:hs-p-overlap}
\end{figure*}
\item
Because HS bands are very narrow, on the order of 10 nm, large ground cells must be used to harvest sufficient energy, typically 30m or more.
Nonetheless, the acquired data are rather noisy, and some bands or groups of bands must be discarded immediately because they are affected by acquisition errors
and even the remaining ones may not have constant quality.
Furthermore, there is a large resolution mismatch with the PAN, say, 30m vs. 5m,
with a large amount of missing information that the pansharpening tool is called upon to recover.
\item
HS images are huge.
Even assuming that complexity grows linearly with the number of bands, the computational burden of pansharpening is much larger than with MS images.
Moreover, solutions based on machine learning require comparably large volumes of data for training, which may simply not be available.
\end{itemize}
%\red{[MORTI E FERITI]
% i) optimization schedules become critical issues \cite{Guarino2023}.
%ii) employing methods originally designed for MS pansharpening introduces spectral distortion, because of features which may not be visible in both PAN and individual HS spectral bands.
%}

The first HS pansharpening methods proposed in the literature,
beginning with the 2007 paper by Aiazzi {\it et al.} \cite{Capo07},
were indeed ingenious adaptations of solutions originally conceived for the MS case.
In the following years, many more proposals considered the same path,
with approaches based on Bayesian models \cite{Zhang2009, Simoes2015, Wei2015, Wei2015a},
matrix factorization \cite{Berne2010, Moel09, Huang2013},
variational optimization \cite{Kawakami2011},
component substitution and multiresolution analysis \cite{Vivone2014b}
(see also the 2015 review by Loncan {\it et al.} \cite{Loncan2015}).
Some methods specifically designed for HS pansharpening were also proposed,
based on guided filtering \cite{Qu17}, variational optimization \cite{Adde17,Huan17}, saliency-based component substitution \cite{Dong20}.

In the meantime, however, the deep learning (DL) revolution had already hit the field of remote sensing.
Starting from the 2016 seminal work by Masi {\it et al.} \cite{Masi2016},
deep learning is by now a {\em de-facto} standard for both MS \cite{Yang2017, Deng2022, Ciotola2023a} and HS \cite{He2019a, Zheng2020,Bandara2022,Guarino2023} pansharpening.
Limiting the analysis to the latter case,
in \cite{He2019a} and \cite{He2020} dedicated convolutional neural networks (CNNs) were designed to strengthen the spectral prediction capability.
In \cite{Zheng2020}, instead, a dual-attention residual network was proposed, with a deep HS image prior module for HS upsampling.
In \cite{He2022, He2022a} the problem of scale and resolution variability was tackled, by means of a CNN with arbitrary-scale attention modules \cite{He2022a}.
An overcomplete residual network that learns high-level features using constrained receptive fields was proposed in \cite{Bandara2022}.
Multibranch network architectures were also explored in several recent works \cite{Guan2022, Wu22, Qu22a}.
Other methods adopted classical pansharpening paradigms and used DL modules to estimate the model parameters \cite{Dong22} or to process the extracted features \cite{Dong22a}.
All the above methods use supervised learning and perform training at reduced resolution using a subsampled version of the original data as ground truth (GT).
Recently, to avoid the quality loss induced by subsampling, a paradigm shift towards full-resolution training is taking place
for both MS \cite{Seo2020, Ciotola2022, Ciotola2023} and HS \cite{Nie2022,Guarino2023, Guarino2023a} pansharpening,
with novel unsupervised learning strategies and suitably defined loss functions.

This quick analysis shows growing research activity on HS pansharpening, with contributions primarily focused on DL-based methods.
Performance, however, does not appear to be improving at a comparable rate, and the success seen in related fields is only coming slowly.
A first reason for this failure is the scarcity and excessive heterogeneity of the data available to the scientific community, a well-known plague of remote sensing.
Limiting the scope to the DL-oriented papers mentioned before, 
we observe that different research groups work on datasets that are small and incompatible with one another,
acquired by different sensors, characterized by different spectral ranges, with different numbers of bands and different PAN-HS resolution ratios.
A further problem is that the experimental protocols and metrics themselves used for performance evaluation often vary as well.
Things are even worse for DL-based methods; virtually all new proposals these days need to be trained on large, high-quality datasets to express their full potential.
Under these conditions, a new proposal can hardly be compared to previous ones based on numerical performance indicators, nor will it work on different data, unless properly adapted and retrained.
All these problems represent major obstacles to scientific progress in this field.

This work tackles the above issues by providing a reliable framework that will help develop and assess new solutions for HS pansharpening.
We begin by scanning the literature to identify a set of benchmark state-of-the-art (SoTA) methods.
Our analysis follows an operational approach,
focusing on methods that provide promising experimental results in some respect and can be faithfully replicated.
After this phase, we build a new development and testing toolbox.
We first assemble\footnote{We use the PRISMA images, 
which are proprietary but can be downloaded for free with permission from the Italian Space Agency.} a large dataset of high-quality PAN+HS images 
to enable the reliable and uniform performance assessment of all methods and to support the accurate training of DL-based methods.
Then, we re-implement all the selected methods in a unique Python-based framework.
This is achieved by either using the original codes as baseline, when available,
or following the authors descriptions and guidelines.
Among these, the DL-based ones are all retrained on the same suitably designed wide and rich dataset.
Finally, we extensively compare the selected methods on the test datasets at both original resolution (or {\em full} resolution, FR),
with no-reference quality indexes due to the lack of ground truth, and in a {\em reduced} resolution (RR) space using reference-based accuracy indexes
that quantify the error directly. 
The accuracy indexes, selected among the most credited ones in the literature, are also provided as part of the toolbox.
Hereinafter, the terms FR and RR data are always referred to the input given to the pansharpening process
and can also be also termed {\em real} and {\em simulated} data, respectively, 
as the latter indicates data obtained through a suitable degradation processing of the former.
%Then, we re-implement, \red{either using the available code or following the guidelines outlined in the papers,} 
%all the selected methods in Python 
%(retraining the DL models \red{on the same, proposed, dataset), to ensure seamless integration within the proposed framework,} 
%and assess their performance both at reduced resolution and full resolution using the most accredited metrics in the literature.
Eventually,
we provide the community with an easy-to-use toolbox comprising data, tools, and results that we hope
will simplify the development of new solutions and drive progress in this field.
As far as we know, 
this is the first attempt to provide a critical comparison for the HS pansharpening problem. 
A unique instance of review fully devoted to HS pansharpening is in \cite{Loncan2015} but referring to dated approaches without involving DL solutions. Moreover, in \cite{Loncan2015}, the assessment has been performed using simulated data (including simulated PANs) and just at reduced resolution. Instead, in this work, real data (both HS and PAN) are exploited, and the validation relies upon both reduced resolution and full resolution procedures.

According to this plan, the remainder of the paper is organized as follows:
Section~\ref{sec:methods} provides a review of HS pansharpening solutions, with a special focus on those more amenable to be implemented and assessed;
Section~\ref{sec:metrics} describes the proposed approach to quality assessment;
Section~\ref{sec:dataset_analysis} reviews the datasets used so far in HS pansharpening and describes the newly proposed dataset;
Section~\ref{sec:exp} presents and discusses numerical and visual experimental results both at reduced and full resolution, extracting general guidelines for future work.
Finally Section~\ref{sec:conclusions} draws conclusions and outlines future work.

\section{Methods}
\label{sec:methods}

\begin{table}
\caption{Benchmark HS pansharpening methods.}
\footnotesize
%\scriptsize
\centering
\setlength{\tabcolsep}{2pt}
\begin{tabular}{lcp{5.5cm}}%{p{2cm}p{3cm}p{1.5cm}}%{p{8.4cm}}
\hline
\bf \ru Name & \bf Ref  & \bf Summary\\
\hline
\hline
\ru EXP                 &    & Approximation of the ideal interpolator \\
\hline
\multicolumn{3}{c}{\bf \ru Component Substitution (CS) methods} \\[1mm]
%GS                 & \cite{Laben2000}    & Gram-Schmidt \\
GSA                 & \cite{Aiazzi2007}   & Gram-Schmidt adaptive component substitution \\
BT-H                & \cite{Lolli2017}    & Brovey transform with haze correction \\
%BDSD               & \cite{Garzelli2008} & Band-dependent spatial detail injection\\
BDSD-PC             & \cite{Vivone2019}   & Band-dependent spatial detail injection with physical constraint\\
PRACS               & \cite{Choi2011}     & Partial replacement adaptive CS \\ \hline
\multicolumn{3}{c}{\bf \ru Multiresolution Analysis (MRA) methods} \\[1mm]
%MTF-GLP            & \cite{Aiazzi2006}   & MTF-matched Generalized Laplacian Pyramid \\
MTF-GLP-FS          & \cite{Vivone2018a}  & Modulation Transfer Function (MTF)-matched Generalized Laplacian Pyramid (MTF-GLP) with fusion rule at full scale \\
MTF-GLP-HPM         & \cite{Alparone2017} & MTF-GLP with high pass modulation\\
% MTF-GLP-HPM-H     &  \cite{Lolli2017}   & MTF-GLP-HPM with haze correction\\
MTF-GLP-HPM-R       & \cite{Vivone2017}   & MTF-GLP-HPM with regression-based spectral matching\\
AWLP                & \cite{Otazu2005}    & Additive wavelet luminance proportional \\
MF                  & \cite{Restaino2016} & Nonlinear decomposition with morphological filters \\ \hline
\multicolumn{3}{c}{\bf \ru Model-Based Optimization (MBO) methods}\\[1mm]
% Bayesian Naive    & \cite{Wei2015b}     & Bayesian estimation with naive independent Gaussian prior \\ %\cite{Wei201},
% Bayesian Sparse   & \cite{Wei2015a}     & Bayesian estimation with sparse representations-based prior \\
HySURE              & \cite{Simoes2015}   & Bayesian estimation with vector total variation prior \\
SR-D                & \cite{Vicinanza2015}& Sparse representations-based detail injection \\
TV                  & \cite{Palsson2014}  & Total variation-based pansharpening \\ \hline
\multicolumn{3}{c}{\bf \ru Deep Learning (DL) methods} \\[1mm]
HyperPNN            & \cite{He2019a}      & 7-layer net with spectral encoder-decoder structure\\
HSpeNet             & \cite{He2020}       & Advanced version of HyperPNN, with deeper architecture and Spectral Angle Mapper (SAM) loss term\\
DHP-DARN            & \cite{Zheng2020}    & Deep residual channel-spatial attention net with Deep Image Prior (DIP) for HS resizing\\
DIP-HyperKite       &  \cite{Bandara2022} & Overcomplete net with DIP for HS resizing\\
HyperDSNet          & \cite{Zhuo2022}     & Spectral attention-based detail injection from deep-shallow features \\
R-PNN (unsup.)      & \cite{Guarino2023}  & Band-wise pansharpening using modified {Z-PNN} \cite{Ciotola2022} with tuning propagation \\
PCA-Z-PNN (unsup.)  & \cite{Guarino2023a} & Z-PNN model with PCA-based input reduction\\
\hline
\end{tabular}
\label{tab:methods}
\end{table}

\renewcommand{\H}{\mathbf{H}}
\newcommand{\X}{\mathbf{X}}
\renewcommand{\P}{\mathbf{P}}
\newcommand{\Q}{\mathbf{Q}}
\newcommand{\C}{\mathbf{I}}
\newcommand{\wt}[1]{\widetilde{\mathbf{#1}}}
\newcommand{\wh}[1]{\widehat{\mathbf{#1}}}

\begin{table}
\caption{Main symbols used in the paper.}
\footnotesize
\centering
\setlength{\tabcolsep}{2pt}
\begin{tabular}{cll}%{p{2cm}p{3cm}p{1.5cm}}%{p{8.4cm}}
\hline
\textbf{\ru Symbol} & \textbf{Dimensions}    & \textbf{Meaning}\\\hline
%\red{HS} && Hyperspectral \\
%\red{PAN} & & Panchromatic \\
$R$                  & Scalar \ru                 & Resolution ratio \\
$B$                  & Scalar                     & \# of HS bands \\
$N$                  & Scalar                     & \# of patches \\
$w, h$               & Scalars                    & HS width and height, respectively \\
$W, H$               & Scalars                    & PAN width and height, respectively \\
$\H$                 & $[w{\times}h{\times}B]$    & Original HS image \\
$\P$                 & $[W{\times}H]$             & Original PAN image \\
$\wt{H}$             & $[W{\times}H{\times}B]$    & $R{\times}R$ upsampling of $\H$ \\
$\wh{H}$             & $[W{\times}H{\times}B]$    & Pansharpening of $\H$ \\
$\wt{X}$ or $\wh{X}$ & $[W{\times}H{\times}\ast]$ & Upsampling or pansharpening of any $\X$ \\
\hline
\end{tabular}
\label{tab:symbols}
\end{table}

This section reviews SoTA solutions for HS pansharpening.
Though described or mentioned, not all reviewed methods are actually part of the experimental framework.
Our critical review, in fact, aims to achieve a fair comparison where the involved methods are developed in the same framework
and trained/optimized on the same datasets to prevent experimental biases.
On the other hand, 
an overall limit to the number of solutions to develop and assess is also necessary to achieve a handy and trustworthy benchmark.
Eventually, the experimental part will focus on a subset of representative methods, 
selected because they:
a) are distinctive from the methodological point of view;
b) are competitive in terms of quality and/or computational efficiency;
c) come with a publicly available software code (in any programming language) or are described with sufficient detail to be faithfully coded anew.

The methods actually implemented and assessed, released as part of the proposed benchmarking toolbox,
are gathered in Tab.~\ref{tab:methods}, where they are grouped according to the categories proposed in \cite{Vivone2020}, {\em i.e}:
Component Substitution (CS),
Multi-Resolution Analysis (MRA),
Model-Based Optimization (MBO) which extends the usual Variational Optimization class,
and DL.
In the pansharpening literature, in addition to the fusion methods it is also custom 
to provide the simple expansion (EXP) of the HS (with no fusion with the PAN) obtained by
an approximation of the ideal interpolator. 
This because such a product is a latent variable of many methods and represents 
a spectral benchmark. 
For the sake of readability, 
the following sectioning matches the categories given in Tab.~\ref{tab:methods}
and paragraphs correspond to the associated methods, 
though variants or alternative solutions out of benchmark are also mentioned or briefly detailed.
In the upcoming description, we will use the symbols recalled in Tab.~\ref{tab:symbols}.

\subsection{Component substitution}

%The

CS methods are based on the use of a suitable transformation
that brings the data to a domain where spatial and spectral components can be easily separated.
Then, the spatial component is replaced by the available PAN and the data are transformed back into the original domain.
By using a linear transformation and considering the substitution of a single component, a fast pansharpening process is obtained \cite{Tu2001}.
These techniques were initially proposed for MS pansharpening and, subsequently, extended to the HS case.
Preliminarily, the HS datacube $\H\in\mathbb{R}^{w{\times}h{\times}B}$ is spatially resized to match the target pansharpening scale
using (a polynomial approximation of) the ideal interpolator \cite{Vivone2014},
yielding $\wt{H}\in\mathbb{R}^{W{\times}H{\times}B}$.
The component to be replaced, $\C\in \mathbb{R}^{W{\times}H}$, is obtained through a simple linear combination of the upsampled bands $\{\H^b\}$ of $\wt{H}$,
through suitably defined weights $\mathbf{w}=[w_1, \ldots, w_B]^T$:
\begin{equation}
    \C = \sum_{b=1}^B w_b \wt{H}^b.
    \label{eq:component}
\end{equation}
Both $\P$ and $\C$ are defined in the panchromatic domain and their difference, after histogram equalization,
represents the high-frequency details missing in the original image.
Therefore, the pansharpened bands $\{\wh{H}^b\}$ of $\wh{H}\in\mathbb{R}^{W{\times}H{\times}B}$
are obtained by injecting these details in the upsampled HS bands, weighted by suitable coefficients, the injection gains $\mathbf{g}=[g_1, \ldots, g_B]$
\begin{equation}
    \wh{H}^b = \wt{H}^b + g_b(\P-\C), \;\;\; b=1, \ldots, B,
    \label{eq:cs}
\end{equation}
By changing the transformation, different combinations of injection gains and weights are needed.

\subsubsection{GSA}
A powerful example of the CS approach is based on the orthogonal Gram-Schmidt (GS) decomposition of $\H$ \cite{Laben2000}.
Different variants can be obtained by changing the definition of the first component of the decomposition, the one to be replaced.
The basic implementation proposed in \cite{Laben2000} employs a simple uniform average: $w_b=1/B,\, \forall b$.
The component substitution is then completed by inverting the orthogonal decomposition using the injection gains:
\begin{equation}
g_b = \frac{{\rm Cov}(\wt{H}^b,\C)}{{\rm Var}(\C)},\;\; b=1,\ldots,B,
\label{eq:GS gains}
\end{equation}
where ${\rm Cov}(\cdot)$ and ${\rm Var}(\cdot)$ indicate the covariance and variance operators, respectively.
This solution, however, does not take into account the different levels of correlation between $\P$ and each individual band of $\H$, giving rise to heavy spectral distortion.
For this reason, a more robust variant, known as Gram-Schmidt Adaptive (GSA),
was proposed in \cite{Aiazzi2006} for MS pansharpening and later applied to the HS case \cite{Vivone2014b}.
The latter is the first method selected for our experimental assessment.
GSA differs from the classical GS implementation in that the weights $\{w_b\}$ used to compute $\C$ are not all equal but are
estimated by minimizing the mean square error between $\C$ and a spatially degraded version of the PAN image.

\subsubsection{BT-H}
Another CS method considered in this work is BT-H \cite{Lolli2017},
which relies on an improved version of the Brovey Transform (BT) \cite{Gillespie1987} that accounts for haze.
In this case, space-variant injection gains are used, defined as
\begin{equation}
    \mathbf{G}^b_n = \frac{\wt{H}^b_n-h^b}{\C_n-h}, \;\;\; \forall\, n, %\mbox{pixel locations}\,n
    \label{eq:BT-H}
\end{equation}
where the ratio is meant to be pixel-wise, $n$ indicates the pixel location, $h^b$ the haze in the $b$-th HS band and $h$ is the haze in both $\P$ and $\C$.
Of course, the product between the gains and $(\P-\C)$ in Eq.~\ref{eq:cs} must be pixel-wise as well.
The weights $\mathbf{w}$ are estimated by minimizing the mean square error between $\C$ and a reduced resolution version of $\P$.

\subsubsection{BDSD-PC}
In \cite{Garzelli2008} both weights $\mathbf{w}$ and gains $\mathbf{g}$ are obtained by minimizing
the mean squared distance between the fused image given by Eq.~\ref{eq:cs} and the reference (GT) pansharpened image.
Since the latter is not available, the optimization problem is shifted to the reduced resolution domain by means of a suitable scaling of the input images.
This approach, known as Band-Dependent Spatial-Detail (BDSD) injection method \cite{Garzelli2008},
is here considered in the version proposed in \cite{Vivone2019} where a Physical Constraint is introduced (BDSD-PC) to regularize the estimation of the coefficients.

\subsubsection{PRACS}
The last CS solution considered in our study is the Partial Replacement Adaptive Component Substitution (PRACS) \cite{Choi2011},
where the replacement of $\C$ is done, band-wise, with a suited weighted sum of $\P$ and $\wt{H}^b$ rather than just using $\P$.

In general, CS methods are characterized by high spatial fidelity, low processing time, and robustness to registration errors and aliasing \cite{Baronti2011,Vivone2014}.
On the downside, they tend to introduce significant spectral distortion due to the spectral mismatch between $\P$ and $\H$ \cite{Thomas2008}.
This becomes a rather serious issue in the HS case, as opposed to the MS case,
because of the large number of spectral bands that present little or no correlation with the PAN.

\subsection{Multiresolution analysis}
Methods based on Multi-Resolution Analysis (MRA) \cite{Vivone2014,Vivone2020,Loncan2015} are formally very similar to CS methods,
as they also involve the injection of spatial details in HS upsampled bands:
\begin{equation}
    \wh{H}^b = \wt{H}^b + \mathbf{G}^b\cdot(\P-\P^{\rm lp}), \;\;\; b=1, \ldots, B,
    \label{eq:mra}
\end{equation}
The fundamental difference is in how these spatial details, or so-called PAN details, are extracted.
In MRA, they are computed as the difference between $\P$ and its low-pass filtered version, $\P^{\rm lp}$,
while in CS as the difference between $\P$ and a weighted average of $\wt{H}$ along the spectral dimension.
The use of different filtering strategies and injection rules gives rise to different MRA solutions.
SoTA injection schemes follow two approaches \cite{Vivone2013},
commonly referred to as {\em projective} and {\em multiplicative} or high pass modulation (HPM).
The former uses spatially constant gains $\mathbf{G}^b$,
while the latter allows them to vary spatially so that the injection of detail can be modulated in the spatial domain.
In more detail, the HPM injection gains are defined as $\mathbf{G}^b_n = \wt{H}^b_n/\P^{\rm lp}_n, \forall n$,
with the goal of reproducing the local intensity contrast of the PAN image \cite{Vivone2013}.
Different MRA techniques, however, are characterized mainly by how the PAN image is low-pass filtered.

\subsubsection{Laplacian-based techniques: MTF-GLP-*}
A popular option is to use a Gaussian filter matched with the Modulation Transfer Function (MTF) of the high spectral resolution sensor, usually designed based on information provided by the manufacturer, such as the gains at the Nyquist frequency.
In the context of multiresolution decomposition, this approach is usually called Generalized Laplacian Pyramid (GLP), since the desired detail $(\P-\P^{\rm lp})$ is approximated by Laplacian filtering.

In this work, we consider several SoTA techniques belonging to this category.
The MTF-GLP-FS method is based on the Full Scale (FS) projective injection rule proposed in \cite{Vivone2018a} working directly at full resolution.
It has proven especially valuable in fusion problems characterized by a large resolution ratio.
The classical multiplicative injection model is considered in MTF-GLP-HPM \cite{Vivone2014b}, implemented with suitable histogram matching to account for possible spectral distortions.
Another variant of the multiplicative injection rule is introduced in MTF-GLP-HPM-R \cite{Vivone2017}.
Here, the ``R'' stands for regression, because the spectral matching between each $\wt{H}^b$ and $\P^{\rm lp}$ is achieved employing multivariate linear regression,
better motivated than the classical histogram matching procedure under a physical point of view \cite{Vivone2017}.

\subsubsection{AWLP}
Other solutions are based on the wavelet transform.
One of the most popular is the {\em Additive Wavelet Luminance Proportional} (AWLP) method \cite{Otazu2005} where PAN detail extraction relies on undecimated {\em {\`a} trous} wavelet transform.
This method also uses a multiplicative injection scheme and histogram equalization.

\subsubsection{MF}
An example of non-linear decomposition, called Morphological Filters (MF), is instead explored in \cite{Restaino2016}, where the pyramid decomposition is built using morphological half-gradient filters.

Overall, the main advantages of MRA approaches are temporal coherence \cite{Vivone2020},
spectral consistency \cite{Vivone2014}, and robustness to aliasing under proper conditions \cite{Baronti2011}.
On the downside, they are more sensible than CS methods to mis-registration and spatial distortion
and are more computationally demanding.

\subsection{Model-based optimization}
Another important category includes methods based on an explicit analytic model of the problem, to be solved through suitable optimization techniques.
These are further divided into three subcategories:
Bayesian \cite{Wei2015b, Wei2015a, Simoes2015},
dictionary-based (namely, based on sparse representations) \cite{Vicinanza2015}, and
variational \cite{Palsson2014}.

\subsubsection{HySURE (a Bayesian approach)}
A mathematically appealing way to address pansharpening is to cast it as an inverse problem in a probabilistic framework, to be solved by means of Bayesian estimation.
Preliminarily, following a methodology well-known in linear HS unmixing \cite{Bioucas2012},
the target high-resolution HS image is assumed to belong to a low-dimensional subspace,
such to be expressed as
\begin{equation}
    \wh{H} = \mathbf{M} \wh{X},
    \label{eq:subspace}
\end{equation}
where
$\wh{H}$ has been arranged in a $B{\times}HW$ 2-D matrix by collapsing the two spatial dimensions,
$\wh{X}$ is a $C{\times}HW$ 2-D matrix, with $C \ll B$, projection of $\wh{H}$ in the low-dimensional subspace,
and $\mathbf{M}$ is the $B{\times}C$ projection matrix, whose columns are the basis of the new subspace.
The transformation $\mathbf{M}$ can be obtained via different approaches, {\it e.g.}, principal component analysis \cite{Farrell2005} or vertex component analysis \cite{Nascimento2005}.
Once a solution, $\wh{X}$, is found, it can be re-mapped into the original HS space through Eq.~\ref{eq:subspace}.
The probabilistic modelling of the problem requires two items:
a {\em likelihood} term, $p(\P, \H|\wh{X})$ relating the observations, $\H$ and $\P$, to the reduced-rank representation $\wh{X}$,
and a {\em prior} term, $p(\wh{X})$.
Then, the pansharpened image, $\wh{X}_{\rm MAP}$, can be found according to the maximum {\em a posteriori} (MAP) criterion as:
\begin{equation}
    \wh{X}_{\rm MAP} = \underset{\wh{X}}{\rm argmax}\, p(\P, \H|\wh{X}) p(\wh{X}).
    \label{eq:bayes}
\end{equation}
To solve the problem, some reasonable simplifying hypotheses can be made \cite{Molina1999, Hardie2004, Molina2008}:
\begin{eqnarray}
    \P &=& \phi_{\rm P}(\mathbf{M}\wh{X}) +  \mathbf{N}_{\rm P},\\
    \H &=& \phi_{\rm H}(\mathbf{M}\wh{X}) +  \mathbf{N}_{\rm H},
\end{eqnarray}
where
$\phi_{\rm P}(\cdot)$ denotes a linear mapping between $\wh{H}$ and $\P$,
$\phi_{\rm H}(\cdot)$ is the spatial degradation model (low-pass filtering and decimation) relating high and low-resolution HS images,
$\mathbf{N}_{\rm P}$ and $\mathbf{N}_{\rm H}$ are zero-mean normal-distributed random matrices \cite{Gupta2000}.
Assuming the conditional independence of the two noise terms, the problem simplifies further to
\begin{equation}
    \wh{X}_{\rm MAP} = \underset{\wh{X}}{\rm argmax}\, p(\P|\wh{X}) p(\H|\wh{X}) p(\wh{X}).
\end{equation}
%where
%\begin{eqnarray}
%&&\P|\wh{X} \sim \mathcal{MN}(\phi_{\rm P}(\mathbf{M}\wh{X}),\sigma_{\rm P}^2,\mathbf{I}), \nonumber \\
%&&\H|\wh{X} \sim \mathcal{MN}(\phi_{\rm H}(\mathbf{M}\wh{X}),\boldsymbol\Sigma_{\rm H},\mathbf{I}), \nonumber
%\end{eqnarray}
%being $\mathcal{MN}(\cdot)$ the multivariate normal distribution, $\sigma_{\rm P}$ and $\boldsymbol\Sigma_{\rm H}$ the variance and covariance of $\P$ and $\H$, respectively,
%and $\mathbf{I}$ an identity matrix of appropriate size to represent the model independence in the spatial dimension \cite{Gupta2000}.
To proceed to the optimization phase the prior distribution $p(\wh{X})$ must be set.
In \cite{Wei2015b} a simple pixel-wise independent Gaussian prior is assumed,
while a more complex prior based on sparse representation is considered in \cite{Wei2015a}.
However, in both cases the resulting HS pansharpening algorithms provide unsatisfactory results, and hence they are not further considered here.
On the contrary,
the Hyperspectral SUperREsolution (HySURE) method proposed in \cite{Simoes2015} is among the most promising in the field,
proving clearly superior to classical Bayesian solutions in our experiments.
It is characterized by an edge-preserving regularizing prior, a form of vector total variation,
whose objective is to promote piecewise-smooth solutions with discontinuity aligned across the HS bands.
To limit complexity,
optimization is pursued in a low-dimensional subspace through the Alternating Direction Method of Multipliers (ADMM).

\subsubsection{SR-D}
To represent dictionary-based methods, we selected the Sparse Representation of Details (SR-D) proposed in \cite{Vicinanza2015}.
With this approach, the spatial details to inject into $\wt{H}$ are built from a suitably learned dictionary of patches.
Let $\mathbf{D}^{\rm h}$ and $\mathbf{D}^{\rm l}$ be two paired dictionaries of patches at high resolution and low resolution, respectively.
With these data, we aim at estimating a desired high-resolution patch, $\mathbf{y}^{\rm h}$, based on its low-resolution version, $\mathbf{y}^{\rm l}$.
In detail, we solve the following problem
\begin{equation}
\hat{\boldsymbol\alpha} = \arg\min \|\boldsymbol\alpha \|_0 \;\;\; {\rm subject\;to} \;\;\; \mathbf{y}^{\rm l} = \mathbf{D}^{\rm l}\boldsymbol\alpha,
\label{eq:alpha}
\end{equation}
where $\|\cdot\|_0$ is the $\ell_0$ pseudonorm, used to induce sparsity in the solution.
In practice, the target patch is approximated by a linear combination of patches in the low-resolution dictionary, with the constraint to use the least possible patches.
Once the weights of the linear combination are obtained,
they are used to estimate the high-resolution target patch by using the very same linear combination of the paired high-resolution patches in $\mathbf{D}^{\rm h}$.
Eventually, we have that $\mathbf{y}^{\rm h} = \mathbf{D}^{\rm h}\boldsymbol\alpha$, where $\mathbf{y}^{\rm h}$ is the pansharpened product in vector form.
Of course, the whole method relies on a strong assumption of invariance across scales (see \cite{Vicinanza2015} for more details).

\subsubsection{TV}
The variational optimization method proposed in \cite{Palsson2014}, simply referred to as Total Variation (TV) here,
is defined by the following TV-regularized least squares problem:
\begin{equation}
    \| \mathbf{y} - \mathbf{M}\wh{X}  \|_2 + \lambda {\rm TV}(\wh{X}),
\end{equation}
where
$\mathbf{y}$ is a suitably reshaped composition of the HS and PAN components,
$\mathbf{M} = [\mathbf{M}_1^T, \mathbf{M}_2^T]$ consists of a decimation matrix $\mathbf{M}_1$ and a weight matrix $\mathbf{M}_2$ summarizing the (supposed) linear dependence between HS and PAN,
${\rm TV}(\cdot)$ is an isotropic TV regularizer, and
$\lambda$ is a balance parameter.
The pansharpened image $\mathbf{x}$ is obtained by minimizing the above convex cost function using a majorization–minimization algorithm detailed in \cite{Palsson2014}.

\subsection{Deep learning}
As already said, DL is by far the most popular approach for MS and HS pansharpening, these days.
Following our criteria,
we have selected seven SoTA methods for our toolbox, five of them based on supervised learning \cite{He2019a,He2020,Zheng2020,Bandara2022, Zhuo2022}, and two \cite{Guarino2023,Guarino2023a} on unsupervised learning.
%\red{
%In this review, we focused on nine SoTA methods, including six based on supervised learning \cite{He2019a,He2020,Zheng2020,Bandara2022,Bandara2022a,Zhuo2022} and three utilizing unsupervised learning approaches \cite{Nie2022,Guarino2023,Guarino2023a}. 
%}
%From these, we selected seven SoTA methods for our toolbox, five of them based on supervised learning \cite{He2019a,He2020,Zheng2020,Bandara2022, Zhuo2022}, and two \cite{Guarino2023,Guarino2023a} on unsupervised learning.
Due to the lack of GT, supervised DL models are trained following the classic Wald's protocol \cite{Wald1997}, as suggested in \cite{Masi2016}.
%originally introduced for the purpose of quality assessment in pansharpening.
This consists of low-pass filtering and decimating (in both directions) both PAN and HS by the same factor $R$, equal to the PAN-MS resolution ratio.
These downsampled images are then used as synthetic PAN and HS input for training the pansharpening network, using the original HS data as GT.
The network trained on reduced resolution data is then used to pansharpen full resolution data, relying on a scale invariance assumption.
This latter hypothesis, however, is rather shaky, which is why unsupervised DL methods are appearing with increasing frequency lately.

\subsubsection{HyperPNN}
This model, proposed in \cite{He2019a},
has a relatively shallow 7-layer architecture, comprising a spectral encoder, a spatial-spectral inference subnet, and a spectral decoder.
More precisely,
the network takes in input the interpolated HS image $\wt{H}$ and the PAN $\P$,
yielding in output the pansharpened image $\wh{H}$ by composing the following three subnets:
%{\small
\begin{eqnarray}
\hspace{-5mm}&\hspace{-5mm}&e:\; \wt{H}\in\mathbb{R}^{W{\times}H{\times}B} \longrightarrow  \wt{X} = e(\wt{H}) \in\mathbb{R}^{W{\times}H{\times}64}, \nonumber \\
\hspace{-5mm}&\hspace{-5mm}&f:\; (\wt{X},\P)\in\mathbb{R}^{W{\times}H{\times}65} \longrightarrow  \wh{Y} = f(\wt{X},\P) \in\mathbb{R}^{W{\times}H{\times}64}, \nonumber \\
\hspace{-5mm}&\hspace{-5mm}&d:\; \wh{Y}\in\mathbb{R}^{W{\times}H{\times}64} \longrightarrow  \wh{H} = d(\wh{Y}) \in\mathbb{R}^{W{\times}H{\times}B}. \nonumber
\end{eqnarray}
%\normalsize}
Both encoder ($e$) and decoder ($d$) work exclusively in the spectral domain, with 1$\times$1 convolutions (two layers each), and are responsible for spectral preservation.
The middle subnet, $f$, works jointly in the spatial and spectral domains with three convolutional layers, each with a 3$\times$3 receptive field and 64 output features.
Two variants are presented in \cite{He2019a}, with or without skip connections over $f$.
Here, we consider only the latter, the most effective,
where the feature volume entering $f$ is brought directly to the output so that the network can focus on reconstructing the residual.
%This is achieved giving to $f$ the form $f(\wt{X},\P) = f_0(\wt{X},\P) + \wt{X}$.}
As loss, the authors use the mean square error (MSE), which is the baseline option to compare a predicted image to the corresponding GT.

It is worth underlining that the HyperPNN network, like other networks described later on, is designed to work with a fixed number of bands.
This is a major limitation in HS pansharpening where the number of available bands changes from sensor to sensor and, due to noise, from image to image.
In fact, to deal with the three images of the dataset (Tab.~\ref{tab:datasets}) the authors had to train three different image-dependent networks.

\subsubsection{HSpeNet}
This model, proposed in \cite{He2020}, improves upon HyperPNN in two main aspects: architecture and loss.
The network comprises an additional preprocessing subnet, $g$, that extracts suitable features from the PAN to feed the middle subnet
\begin{eqnarray}
\hspace{-5mm}&\hspace{-5mm}&g:\; \P\in\mathbb{R}^{W{\times}H} \longrightarrow  \Q = g(\P) \in\mathbb{R}^{W{\times}H{\times}16}, \nonumber \\
\hspace{-5mm}&\hspace{-5mm}&f:\; (\wt{X},\Q)\in\mathbb{R}^{W{\times}H{\times}80} \longrightarrow  \wh{Y} = f(\wt{X},\Q) \in\mathbb{R}^{W{\times}H{\times}160}, \nonumber
\end{eqnarray}
%\normalsize
The subnet $g$ comprises two convolutional layers with 3$\times$3 receptive fields and 16 output features each.
The feature fusion subnet $f$ has been upgraded as well to a more effective 5-level DenseNet-like structure.
Finally,
a global skip connection has been introduced yielding an output of subnet $d$ in the form:
\begin{equation}
\wh{H} = d(\wh{Y},\wt{H}) = d_0(\wh{Y}) + \wt{H} \in\mathbb{R}^{W{\times}H{\times}B} \nonumber
\end{equation}
where $d_0$ is a single 1$\times$1 convolutional layer that transforms 160 feature maps in $B$ detail bands
that are added to the smooth component $\wt{H}$ to provide the final pansharpened image, $\wh{H}$.
%This solution is referred to as HSpeNet2 in \cite{He2020} to be distinguished from another slightly different option (HSpeNet1) not considered in this work.
The other important difference with HyperPNN is the loss function,
which includes an additional term based on the Spectral Angle Mapper (SAM) \cite{Yuhas1992} to enforce spectral consistency.

\subsubsection{DHP-DARN}
This method \cite{Zheng2020} relies on two key elements,
the use of a deep hyperspectral prior (DHP) model aimed at improving the preliminary upsampling of $\H$,
and the use of the dual-attention residual network (DARN).
DIP~\cite{Lempitsky2018} are called upon to make up for the scarcity of data typical of many problems.
The idea is that, lacking sufficient training data, the input to the network should be as close as possible to the expected output.
In our case, the input, $\wt{H}$, should be close to the expected result, $\wh{H}$, or, at least, spectrally consistent with the original $\H$.
The overall effect is a sort of prior regularization that prevents possible generalization issues.
Unlike other upsampling options, that work band-wise regardless of spectral dependencies,
the DHP module is tuned online on the very same target image to guarantee that the upsampling $\wt{H}$, when degraded, will return to $\H$.
Then, the actual fusion process is carried out through the main network, denoted by the function $f_{\rm DARN}$,
which consists of a sequence of three subnets by-passed by a global skip connection,
\begin{equation}
    \wh{H} = f_{\rm DARN}(\wt{H},\P) + \wt{H} = \wh{H}^{\rm res} + \wt{H},
\end{equation}
where $\wh{H}^{\rm res}$ is the residue (or detail) component of $\wh{H}$.
The first and the third subnets of $f_{\rm DARN}$ are stacked convolutional blocks while the central section is a sequence residual Channel-Spatial Attention (CSA) modules.
Training is carried out using a $\ell_1$-norm loss function.

%\begin{itemize}
%\item Deep HS prior for resize. Estimated through spectral consistency loss on the target image itself, at full resolution
%\item Channel e spatial attention  resblocks.
%\item Un test (solo visivo) a FR sull'immagine (1/4) dotata di PAN (3x)
%\end{itemize}

\subsubsection{DIP-HyperKite}
Another approach based on the DIP, called DIP-HyperKite, has been proposed in \cite{Bandara2022}.
In this case, the generated prior image $\wt{H}$ is forced to be consistent not only with $\H$ but also with $\P$. 
This is achieved through an additional loss term that compares $\P$ to a weighted average of $\wt{H}$ along the spectral dimension, where the weights are also learned.
A residual learning scheme is used, like in DARN.
However, an innovative architecture is proposed here, 
a sort of inverse U-Net \cite{Ronneberger2015} where pooling and unpooling operations are exchanged, 
with the latter working on the ``encoding'' side and the former moved to the ``decoding'' part.
By doing so, in the central part of the network, the spatial resolution increases up to eight times in both directions compared to the target resolution.
This {\em overcomplete} representation is used because the residue to be predicted, $\wh{H}-\wt{H}$, is mostly concentrated in the higher frequencies.
A spatial expansion enables the frequency domain to extend beyond the limitations imposed by PAN resolution, increasing the network's ability to synthesize high-frequency spatial details.
It goes without saying that the computational complexity increases considerably, both in the training and inference phases.

\subsubsection{HyperDSNet}
This model \cite{Zhuo2022} relies on the use of three key elements, as summarized below:
a set of handcrafted operators $d$ that extract additional differential features from the PAN;
a subnet $f_{\rm DS}$ that extracts multiscale Deep-Shallow (DS) features;
a Spectral Attention (SA) module $f_{\rm SA}$ that generates the output residues $\wh{H}-\wt{H}$ through a suitable combination of the extracted features.

{\small
\begin{eqnarray}
\hspace{-5mm}&\hspace{-5mm}&d:\;          \P \in \mathbb{R}^{W{\times}H} \longrightarrow \mathbf{Q} = d(\P) \in \mathbb{R}^{W{\times}H{\times}6}, \nonumber \\
\hspace{-5mm}&\hspace{-5mm}&f_{\rm DS}:\; (\wt{H},\P,\mathbf{Q}) \in \mathbb{R}^{W{\times}H{\times}(B+7)} \longrightarrow \mathbf{F}^{\rm DS}  \in \mathbb{R}^{W{\times}H{\times}B}, \nonumber \\ %= f_{\rm DS}(\wt{H},\P,\mathbf{Q}) 
\hspace{-5mm}&\hspace{-5mm}&f_{\rm SA}:\; \wt{H} \in \mathbb{R}^{W{\times}H{\times}B} \longrightarrow \mathbf{w}_{\rm SA} = f_{\rm SA}(\wt{H}) \in \mathbb{R}^{1{\times}1{\times}B}, \nonumber \\
\hspace{-5mm}&\hspace{-5mm}&o:\;          \wh{H} = \wt{H} + \wh{H}^{\rm res} = \wt{H} + \mathbf{w}_{\rm SA} \cdot \mathbf{F}^{\rm DS}. \nonumber
\end{eqnarray}
}
The features $\Q$, obtained using classical derivative operators such as Roberts, Prewitt and Sobel, 
feed the subsequent feature extractor $f_{\rm DS}$ together with the input pair $(\wt{H},\P)$.
The DS subnet, composed of a sequence of convolutional layers,
provides output features extracted not only by the last layer but also by intermediate layers (hence deep-shallow),
then reduced to $B$ spectral channels.
In parallel, the SA module computes the weights $\mathbf{w}_{\rm SA}$ 
used eventually in the output block to modulate on a per-band basis the detail injection strength.
The whole network is trained end-to-end according to a traditional supervised scheme using a $\ell_1$-norm loss.

Supervised DL-based methods have great potential, as testified by numerous success stories in closely related fields.
Unfortunately, in the case of HS pansharpening there are many problems that prevent the desired results from being achieved:
\begin{itemize}
\item[(a)] 
    The training is performed on synthetic data, obtained through resolution degradation processes,
	and there is no guarantee that a model trained on such data will work equally well on real full-resolution datasets.
	This is a general limit of any supervised pansharpening network.
\item[(b)] 
    The volume of data available for training, already limited by the lack of freely available HS datasets, is further reduced in the presence of resolution downgrading.
\item[(c)] 
    HS images are characterized by a varying number of spectral bands, due to differences among sensors and also to acquisition noise.
    However, none of the above models can work with a variable number of bands.
    Therefore, even when many images are available, they cannot be joined to form a single, training set and,
    in any case, these methods cannot easily generalize to new images.
\end{itemize}
A partial response to these shortcomings is given by DHP-DARN and DIP-HyperKite, which use DIPs to balance weakly trained networks.
The more natural solution, however, is to use unsupervised training schemes \cite{Seo2020, Luo2020, Ciotola2021, Ciotola2022},
which exploit only original full-resolution data to train the network with no need for GT.
In this case, specific loss functions must be defined and carefully designed to drive the network towards the desired behavior.
These losses comprise at least two terms, referred to as spectral ($\L_\lambda$) and spatial ($\L_S$) consistency loss terms, 
\begin{equation}
    \L = \L_\lambda +\beta \L_S = \L_\lambda(\wh{H}, \H) + \beta \L_S(\wh{H},\P).
\end{equation}
The first term accounts for spectral fidelity and is usually computed by 
projecting $\wh{H}$ into the domain of $\H$ through a resolution downgrading and evaluating their distance by the $\ell_1$ or $\ell_2$ norms.
The second term, responsible for the spatial quality of the fused image, is more difficult to define.
The main options are:
{\it  i)} to synthesize a pseudo-PAN through a weighted average of the bands of $\wh{H}$ and compare it with $\P$;
{\it ii)} to compare each band of $\wh{H}$ individually with $\P$ and then summarize results.
In both cases, the comparison should rely only on the high spatial frequency components of $\wh{H}$.

\subsubsection{R-PNN}
The unsupervised Rolling Pansharpening Neural Network (R-PNN) \cite{Guarino2023} addresses explicitly the issues (a)-(c) mentioned before.
It relies heavily on the target-adaptive strategy, originally introduced for the MS case in both supervised \cite{Scarpa2018} and unsupervised \cite{Ciotola2022} settings,
which consists of fine-tuning the pre-trained network on the target data.
R-PNN uses target adaptation in a ``rolling'' modality, 
that is, spectral bands are pansharpened one at a time, by fine-tuning the network for the current band starting from the weights optimized for the previous one.
Formally, let $\phi_b^{(0)}$ and $\phi_b^{(\infty)}$ be the initial and final (tuned) net parameters for band $b$, 
then $\phi_{b+1}^{(0)}=\phi_{b}^{(\infty)}$ are the initial parameters for band $b+1$,
to be adapted through a number of iterations proportional to the spectral distance between the two bands.
Since adjacent bands are highly correlated, very few tuning iterations are sufficient to obtain accurate results, which limits computational complexity.
In addition,
the pansharpening network is a lightweight residual CNN, called Zoom-PNN (Z-PNN) \cite{Ciotola2022}, adapted to the single-band pansharpening case:

{\small
\begin{equation}
%f:\; (\wt{H}^b,\P;\phi) \in \mathbb{R}^{W{\times}H{\times}2}{\times}\Phi \longrightarrow \wh{H}^b = f(\wt{H}^b,\P;\phi)\in \mathbb{R}^{W{\times}H}
f:\; (\wt{H}^b,\P) \in \mathbb{R}^{W{\times}H{\times}2}{\times}\Phi \longrightarrow \wh{H}^b = f(\wt{H}^b,\P)\in \mathbb{R}^{W{\times}H}
\label{eq:rpnn}
\end{equation}
}
% \small
The unsupervised loss, used both for pre-training and tuning, 
comprises a spectral term $\L_\lambda$ based on the $\ell_1$-norm
and a spatial term $\L_S$ based on the local correlation coefficient \cite{Scarpa2022}.

\subsubsection{PCA-Z-PNN}
A further adaptation of the Z-PNN method \cite{Ciotola2022} to HS pansharpening is proposed in \cite{Guarino2023a} based on PCA.
The key observation is that the hundreds of bands comprising the HS image 
can be transformed into a new space where most of the energy is kept in a much smaller number of components.
PCA is a natural candidate for such transformation and preliminary experiments on typical HS images show that it can compact about 99\% of the energy in just 8 bands,
that is the number of bands used in Z-PNN pansharpening.

With this premise, the method is easily explained.
The input $\wt{H}$ is first whitened using the PCA transform. 
Then, the first 8 principal components $\wt{H}^{\rm PCA}$ are pansharpened using Z-PNN in the target adaptive modality.
Finally, the pansharpened components $\wh{H}^{\rm PCA}$
are concatenated with the remaining low-energy components $\wt{H}^{\rm rem}$ and transformed back to the original space.
In formulas, the process can be summarized as follows:
\begin{eqnarray}
{[\mathbf{W}, \sim, \sim]}              &=& {\rm SVD}(\wt{H}\wt{H}^{T}), \nonumber \\
{[\wt{H}^{\rm PCA},\wt{H}^{\rm rem}]} &=& \wt{H} \mathbf{W}, \nonumber\\
\wh{H}^{\rm PCA}                      &=& f_{\rm Z{-}PNN}(\wt{H}^{\rm PCA},\P), \nonumber \\
\wh{H}                                &=& [\wh{H}^{\rm PCA}, \wt{H}^{\rm rem}] \mathbf{W}^{-1}, \nonumber
%\mathbf{W} &=& {\rm eig}(\wt{H}^{T}\wt{H}), \nonumber \\
%{[\wt{H}^{\rm PCA},\wt{H}^{\rm res}]} &=& \wt{H} \mathbf{W}, \nonumber\\
%\wh{H}^{\rm PCA} &=& f_{\rm Z{-}PNN}(\wt{H}^{\rm PCA},\P), \nonumber \\
%\wh{H} &=& [\wh{H}^{\rm PCA}, \wt{H}^{\rm res}] \mathbf{W}^{-1}, \nonumber
\end{eqnarray}
where $\wt{H}$ is zero-meaned and reshaped as a $WH{\times}B$ matrix,
$\mathbf{W}$ is the $B{\times}B$ matrix whose columns are the eigenvectors of $\wt{H}\wt{H}^{T}$
obtained via SVD decomposition, 
$f_{\rm Z{-}PNN}$ is the pansharpening function,
and $(\cdot)^T$ and $(\cdot)^{-1}$ denote transpose and inverse, respectively.

It is worth underlining that the PCA rotation can change from one image to another with no harm since Z-PNN runs in the target-adaptive modality and adapts to the new statistics.
Along the same line, following experimental evidence,
it has been found effective to pansharp separately the set of bands falling in the visible spectrum and those ranging from near to shortwave infrared, 
applying the above-described scheme twice.

\subsection{Beyond the benchmark}
In addition to the above-reviewed methods, there are recent DL solutions not enclosed in the benchmark that are also worth mentioning.
HyperTransformer \cite{Bandara2022a} is one such method that, 
leveraging recent advances in the computer vision domain, 
involves the use of transformers~\cite{Vaswani2017,Dosovitskiy2020}. 
Specifically, it makes use of a self-attention mechanism
that allows the training of the attention modules by means of a suitable loss that combines a $\ell_1$-norm with a perceptual term \cite{Johnson2016}.
Other transformer-based approaches include MDTP-Net~\cite{Wang2024}, a multi-scale pansharpening network that uses multi-head attention mechanisms and dual Gaussian-Laplacian pyramids, and MFT-GAN~\cite{Shang2024}, which leverages transformers in an unsupervised adversarial training framework.

Instead, DMOEAD \cite{Wu2024} explores multi-task learning in the context of HS pansharpening. In this work a framework is proposed which enhances the performance and generalization of HS pansharpening by incorporating the classification task into the learning process.

Another interesting solution is RE-RANet (Ratio Estimation and Residual Attention Network) \cite{Nie2022},
where the use of a residual attention mechanism is combined with the estimation (via neural networks) of the high-low resolution ratio image.
Such a ratio is estimated over the PAN component and then used as a starting point for the estimation of that of each HS band.
The estimation is performed employing a suitable network trained with an unsupervised loss.

An approach similar to PCA-Z-PNN is instead proposed in \cite{Rui2024} and called PLRDiff.
Likewise PCA-Z-PNN, it leverages the low-rank structure in the spectral domain 
that allows a dimensionality reduction.
Besides, it makes use of a pre-trained diffusion model originally designed for the change detection task \cite{Bandara2022b}
and suitably adapted to the pansharpening of the reduced set of transformed spectral bands.

\section{Quality assessment}
\label{sec:metrics}
The goal of pansharpening is to take two low-quality images, having reduced spatial (the HS component) or spectral resolution (the PAN),
and synthesize a high-quality image at full resolution that cannot be observed in reality.
Since the desired full-resolution image is not observable, there is no GT for objectively measuring the quality of the synthesized image.
As a consequence, assessing the quality of a pansharpening algorithm is by no means trivial and remains essentially an open problem, extensively investigated in the past two decades.

A popular measurement protocol was proposed in \cite{Wald1997}, where fusion products are required to satisfy two properties: {\em consistency} and {\em synthesis}.
The {\em consistency} property states that the pansharpened image, once degraded at the lower spatial resolution of the original HS, should be as similar as possible to the latter.
Similarly, a proper spectral degradation process of the pansharpened image should provide a single-band image as similar as possible to the original PAN.
By this definition, it is clear that consistency can be easily measured.
However, it represents only a check, and even perfect consistency does not ensure that the pansharpened image has the desired quality.
The {\em synthesis} property is more stringent, as it states that the pansharpened image
should be as similar as possible to the HS image that would be acquired by the HS sensor if it had the same (high) spatial resolution as the PAN sensor.
Unfortunately, lacking this latter image, the synthesis property cannot be directly assessed and it has a mostly ideal nature.

One can circumvent the problem by resorting to the so-called Reduced Resolution (RR) assessment \cite{Wald1997,Vivone2021}.
The idea is to run the pansharpening algorithm using as input the spatially downgraded versions of PAN and HS.
The output of the algorithm will have the same resolution as the original HS, which can therefore serve as GT.
Thanks to the presence of a reference image, RR quality assessment is simple and accurate, it only requires a metric for measuring the similarity of multi-band images.
However, there is no guarantee that a method that works well on low-resolution data will work equally well on high-resolution data, namely that a sort of scale-invariance holds.
In addition, the degradation process required by this protocol may introduce biases and errors.
From this point of view, the choice of suitable filters is crucial to ensure the consistency of the pansharpening process.
In particular, before decimating the HS image, filters that match the HS sensor's MTFs should be used \cite{Aiazzi2006}.
For the PAN image, instead, an ideal filter is preferred \cite{Vivone2021}, to preserve the details that would have been seen with a direct RR acquisition.

To overcome these problems, pansharpening products can also be evaluated at full resolution, 
using quality indices specifically developed for this purpose according to the Quality with No Reference (QNR) paradigm~\cite{Alparone2008,Arienzo2022}.
Of course, in the absence of a GT, such quantitative measures remain arbitrary to some extent.
Typically, two complementary quality indexes are considered to measure spatial and spectral consistency.
These may follow opposite trends, with the paradox that the least spectral distortion is obtained when no spatial enhancement is introduced.
Therefore, a suitable combination of them is necessary.

In the rest of this Section, we briefly review the quality indices considered in our toolbox, also summarized in Tab.~\ref{tab:indexes}.
%However, it is worth emphasizing once again that quality assessment in pansharpening is an ill-posed problem, with no simple solution, and is still the subject of intense research.

However, it is worth emphasizing that any quality assessment in pansharpening is an open and still debated problem. 
On the one hand, many researchers~\cite{Luo2020,Uezato2020,Ma2020,Ciotola2023} have questioned the validity of scale invariance, which is a key assumption in low-resolution evaluation. On the other hand, no-reference metrics are often adapted from multispectral imaging, without properly addressing the specific challenges of HS images. Additionally, there is still no widespread agreement on the reliability of these full-resolution metrics~\cite{Alparone2008,Khan2009,Alparone2018,Meng2021,Scarpa2022,Arienzo2022,Guan2023}.

Both the reduced resolution and full resolution approaches have advantages and disadvantages.
A good practice, consistently followed in the literature, is to use a wide range of indices and to always integrate the numerical results with a critical visual inspection of the images by experts.

\begin{table}
\caption{HS pansharpening quality assessment indexes.}
\centering
\footnotesize
\setlength{\tabcolsep}{2pt}
\begin{tabular}{rl} \hline
\multicolumn{2}{c}{\bf \ru RR assessment}\\ \hline
\ru
ERGAS                  & {\em Erreur Relative Globale Adimensionnelle de Synth{\'e}se} \cite{Wald2002} \\
SAM                   & Spectral Angle Mapper \cite{Yuhas1992}  \\ %: average divergence (cosine) between predicted and reference spectral response.\\
%$Q$ \cite{Wang2002}                   & Universal Image Quality Index\\
$Q2^n$             & Multiband extension \cite{Garzelli2009} of Universal Image Quality Index~\cite{Wang2002}\\ \hline \hline
\multicolumn{2}{c}{\bf \ru  FR assessment}\\ \hline
\ru
$\DL$      & Khan's spectral distortion index \cite{Arienzo2022, Khan2009}\\
%$\DS$ \cite{Alparone2008}             & Spatial distortion index\\
$\DS$ & Spatial distortion index \cite{Arienzo2022, Alparone2018}\\
%$\DSR$ \cite{Arienzo2022}             & A variant of $\DS$\\
%$\DR$ \cite{Scarpa2022}               & Correlation distortion index\\
RQNR     & Regression-based QNR index \cite{Arienzo2022, Vivone2023}\\ \hline
\end{tabular}
\label{tab:indexes}
\end{table}

\subsection{RR assessment}
\label{sec:RR}

% \red{TO BE WRITTEN}

%Reduced Resolution (RR) assessment evaluates the degree of similarity between the pansharpened product and an ideal reference, i.e. the original HS image. This procedure can be achieved by degrading the resolutions of both the original HS and PAN and by performing fusion from those degraded data.
%In this context, the choice of the correct filter ensures the consistency of the pansharpening procedure. Thus, it is straightforward the use spatial filters aligned with the Modulation Transfer Function (MTF) of respective sensors.
%Hence, the original HS data play the role of GT on which to measure the similarity with the fused product obtained by combining the degraded versions of the original PAN and HS images.
%The higher the similarity, the better the performance. This similarity degree can be evaluated with different multidimensional score indexes.

Following a common practice for pansharpening assessment \cite{Vivone2020, Vivone2021, Vivone2023},
three well-established reference-based indexes have been implemented and used
to assess the similarity between the fused products and the original HS image playing the role of ground truth.

\subsubsection{ERGAS}
The {\em Erreur Relative Globale Adimensionnelle de Synthèse} (ERGAS) \cite{Wald2002}
assesses the distance between two images by generalizing the concept of root mean square error (RMSE) to the multi-band case,
taking care to normalize, band by band, the radiometric error to the average intensity on the reference image.
In detail, it is defined as follows:
\begin{equation}
{\rm ERGAS}  = \frac{100}{R} \sqrt{\frac{1}{B} \sum_{b=1}^B \left(\frac{{\rm RMSE}_b}{\mu^{\rm GT}_b}\right)^2},
\end{equation}
where
${\rm RMSE}_b$ is the $b$-th band RMSE between predicted and reference images and
$\mu^{\rm GT}_b$ is the average intensity of the $b$-th band of the reference.
ERGAS equals zero if and only if the predicted image is identical to the GT, otherwise it gives a positive error measurement.

\subsubsection{SAM}
The Spectral Angle Mapper (SAM)~\cite{Yuhas1992} quantifies the spectral similarity between prediction and reference image
by measuring the average angle (typically in degrees) between predicted and reference pixel spectral signatures,
$\hat{\mathbf{v}} = \left[\hat{v}_1, \hat{v}_2, \dots, \hat{v}_B \right]$ and $\mathbf{v} = \left[v_1, v_2, \dots, v_B \right]$.
Mathematically, we have:
\begin{equation}
   		{\rm SAM} = {\rm E}
   				\left[ \arccos\left( \frac{\left\langle \mathbf{v},\mathbf{\hat{v}} \right\rangle}{\|\mathbf{v} \|_2 \cdot \| \mathbf{\hat{v}} \|_2}\right)\right],
\end{equation}
where
$\langle \cdot, \cdot \rangle$ indicates the dot product,
$\|\cdot\|_2$ is the $\ell_2$ norm,
$\arccos(\cdot)$ the (positive-valued) arccosine function, and
${\rm E}[\cdot]$ the spatial average.
The optimal value of SAM is zero, obtained for predictions that are pixel-wise proportional to the GT.
Therefore, SAM is invariant to spectral signature scaling, $\hat{\mathbf{v}} = \alpha \mathbf{v}$.

\subsubsection{$Q2^n$} The $Q2^n$ index~\cite{Garzelli2009}
generalizes the single-band Universal Image Quality Index (UIQI)~\cite{Wang2002} to the case of images with multiple bands.
Originally introduced for four spectral bands, it was later expanded to handle images with $2^{n}$ bands~\cite{Garzelli2009}.
Each pixel of a $B$-band image is represented as a hyper-complex number with one real part and $B{-}1$ imaginary parts.
By calling $\bz$ and $\hz$ the hyper-complex representations of a GT pixel and its prediction, respectively,
$Q2^n$ can be expressed as:
\begin{equation}
	Q2^n = {\rm E} \left[ \frac{|\sigma_{\bz,\hz}|}{\sigma_\bz \sigma_\hz}
				\cdot \frac{2 \sigma_\bz \sigma_\hz}{\sigma_\bz^2 + \sigma_\hz^2}
				\cdot \frac{2 \mu_\bz \mu_\hz}{|\mu_\bz|^2+|\mu_\hz|^2} \right],
\end{equation}
where $\sigma_{\cdot, \cdot}$, $\sigma_{\cdot}^2$ and $\mu_\cdot$ indicate covariance, variance and mean for hyper-complex variables, computed on 32$\times$32 patches,
$|\cdot|$ provides the vector module and, again, ${\rm E}[\cdot]$ indicates global spatial average.
The first factor accounts for the correlation between $\bz$ and $\hz$, while the other two measure contrast and intensity biases jointly on all bands.
Unlike ERGAS and SAM, $Q2^n\in[0,1]$ has to be maximized,
and the optimum value (one) is achieved when the first- and second-order statistics of predicted and GT images are equal and, their covariance is maximized.

\subsection{FR assessment}
\label{sec:FR}

Full-resolution assessment typically involves the computation of two distinct and complementary quality indexes \cite{Alparone2008,Arienzo2022,Khan2009,Scarpa2022},
although a few solutions based on other paradigms have also been proposed~\cite{carla2015full,vivone2018bayesian,agudelo2019perceptual}.
Here, we use two well-established indexes for assessing spectral and spatial distortion and hence the consistency with HS and PAN, respectively.
Moreover, we consider a single full-resolution score that summarizes the two previous indexes.

\subsubsection{Khan's spectral distortion index}
$\DL$~\cite{Khan2009} is defined as:
\begin{equation}
    \DL = 1 - Q2^n \left(\wh{H}_{\downarrow},  \H \right),
\end{equation}
where
$\wh{H}_{\downarrow}$ indicates the MTF-based low-pass filtered and decimated version of the pansharpened image $\wh{H}$ while $\H$ is the original HS.
Since the index is defined as $1- Q2^n$, the optimum value is 0 (hence the ``distortion'' name),
and it is obtained when the downsampled version of $\wh{H}$ matches the original HS in terms of first-and second-order statistics in the multiband space.

\subsubsection{Spatial distortion index $\DS$}
This index was proposed in \cite{Alparone2018} to assess the spatial consistency between the fused and PAN images.
It is based on the assumption that the PAN can be expressed as a linear combination of individual spectral bands having the same spatial resolution as the PAN and covering collectively the same bandwidth.
In our context,
this amounts to assuming that $\P$ can be approximated with arbitrary precision by combining the bands of the pansharpened image $\wh{H}$
through suitable weights $\{w_b\}$
\begin{equation}
    \C = \sum_{b=1}^{B} w_{b}\wh{H}^{b},
\end{equation}
In practice, the weights are estimated by minimizing the mean square error between $\C$ and $\P$ and the distortion is given by:
\begin{equation}
    \DS = \frac{\sigma^2_{\C - \P}}{\sigma^2_\P},
\end{equation}
where $\sigma^2_{\C-\P}$ and $\sigma^2_\P$ are the variance of $\C-\P$ and $\P$, respectively, obtained by global averages.
$\DS$ can be interpreted as the fraction of the total variance of $\P$ that cannot be explained by $\C$.
Therefore, the optimal value of $\DS$ is zero, obtained if and only if $\C = \P$.

\subsubsection{Regression-based QNR}
The $\DL$ and $\DS$ indexes measure two different aspects of the quality of pansharpened images.
An algorithm that minimizes one index may cause the other to overgrow, leading to poor overall quality.
Therefore, they should be taken into account jointly, looking for the best trade-off between spatial and spectral distortion.
This is the objective of the Regression-based QNR (RQNR) index defined as follows
(see \cite{Arienzo2022} for additional details):
\begin{equation}
    \mathrm{RQNR} = \left(1-\DL \right)^{\alpha} \left(1-\DS \right)^{\beta}.
\end{equation}
Lacking any specific needs, here we set $\alpha=\beta=1$, as also done in \cite{Vivone2023}.
RQNR reaches its optimal value, 1, when both individual distortion indices vanish, 
and decreases rapidly as even one of them increases.

It is worth to notice the similarity between RQNR and the 
Hybrid Quality with No Reference (HQNR) index~\cite{Aiazzi2014} commonly used in the MS case
which differs from the former for a different formulation of the spatial distortion term $\DS$.
The spatial distortion index in HQNR treats any alterations in the relationship between MS and PAN across different resolution scales as spatial distortions. This approach can lead to coupling effects between the two distortion metrics.
In cases, as it is the case here, where many bands lack spectral overlap with the PAN image, 
spectral distortion becomes dominant and can impact the measurement of spatial distortion in HQNR. 
In contrast, the spatial metric employed in RQNR mitigates these coupling effects
and is therefore preferred \cite{Arienzo2022}.

\section{Benchmarking datasets}
\label{sec:dataset_analysis}

\begin{table*}

\caption{Popular datasets for HS pansharpening.}
%\begin{minipage}{20cm}
\footnotesize
\centering
\setlength{\tabcolsep}{2pt}

\begin{tabular}{l@{\hspace{10mm}}c@{\hspace{5mm}}c@{\hspace{5mm}}c@{\hspace{15mm}}c@{\hspace{5mm}}c@{\hspace{5mm}}}
\hline
\ru \multirow{2}{*}{Dataset Name} & Bands & Spectral & Size  & \multirow{2}{*}{Paper} & \multirow{2}{*}{Repository} \\
\ru & (used/all) & Range & $N\times \left(H \times W\right)$ & & \\[1mm]
\hline
\hline
\ru HYDICE - Washington DC Mall & 191/210 & 0.4--2.4 $\mu$m & 1$\times$(1208$\times$307) & \xmark & \cmark$^a$ \\
MV.C VNIR - Chikusei & 128/512 & 0.4--1.0 $\mu$m & 1$\times$(2517$\times$2335) & \cite{Yokoya2016} & \cmark$^b$\\
ROSIS - Pavia Center & 102/115 & 0.4--0.9 $\mu$m & 1$\times$(1096$\times$715) & \xmark & \cmark$^c$ \\
ROSIS - Pavia University & 103/115 & 0.4--0.9 $\mu$m & 1$\times$(610$\times$340) & \xmark & \cmark$^c$ \\
AVIRIS - Moffet Field & 176/224 & 0.4--2.5 $\mu$m & 1$\times$(395$\times$185) & \xmark  & \cmark$^d$ \\
AVIRIS - Salinas Valley & 204/224 & 0.4--2.5 $\mu$m & 1$\times$(510$\times$215) & \xmark & \cmark$^c$   \\
EO-1 - Botswana & 145/242 & 0.4--2.5 $\mu$m & 1$\times$(1496$\times$256) & \xmark & \cmark$^c$ \\
EO-1/ALI - Halls Creek & 171/242 & 0.4--2.5 $\mu$m & 1$\times$(3483$\times$567) & \xmark & \xmark\\
EO-1/ALI - Los Angeles & 145/242 & 0.4--2.5 $\mu$m & 1$\times$(360$\times$360) & \xmark & \xmark\\
CAVE & 31/31 & 0.4--0.7 $\mu$m & 32$\times$(512$\times$512) & \cite{Yasuma2010} & \cmark$^e$ \\
PRISMA - Barcelona &  59/239 & 0.4--2.5 $\mu$m & 1$\times$(900$\times$900) & \cite{Vivone2023} & \cmark$^f$ \\
PRISMA - Milan & 73/239 & 0.4--2.5 $\mu$m & 1$\times$(900$\times$900) & \cite{Vivone2023} & \cmark$^f$ \\
PRISMA - Prato & 73/239 & 0.4--2.5 $\mu$m & 1$\times$(2400$\times$2400) & \xmark & \xmark \\
PRISMA - Bologna & 69/239 & 0.4--2.5 $\mu$m & 1$\times$(2400$\times$2400) & \cite{Vivone2023} & \cmark$^f$ \\
PRISMA - Florence & 63/239 & 0.4--2.5 $\mu$m & 1$\times$(2400$\times$2400) & \cite{Vivone2023} & \cmark$^f$ \\
\hline
\hline
\multicolumn{6}{l}{\ru $^a$https://engineering.purdue.edu/$\sim$biehl/MultiSpec/hyperspectral.html} \\
\multicolumn{6}{l}{$^b$https://naotoyokoya.com/Download.html} \\
\multicolumn{6}{l}{$^c$https://www.ehu.eus/ccwintco/index.php/Hyperspectral\_Remote\_Sensing\_Scenes} \\
\multicolumn{6}{l}{$^d$https://aviris.jpl.nasa.gov/data/free\_data.html } \\
\multicolumn{6}{l}{$^e$https://cave.cs.columbia.edu/projects/categories/project?cid=Computational+Imaging\&pid=} \\
\multicolumn{6}{l}{\hspace{0.2em} Generalized+Assorted+Pixel+Camera} \\
\multicolumn{6}{l}{$^f$https://openremotesensing.net/knowledgebase/panchromatic-and-hyperspectral-image-fusion} \\
\multicolumn{6}{l}{\hspace{0.2em} -outcome-of-the-2022-whispers-hyperspectral-pansharpening-challenge} \\

\hline
\end{tabular}
%\end{minipage}
\label{tab:extra_det}
\end{table*}

\newcommand{\hs}{\hspace{7mm}}
\begin{table*}
\caption{Review of HS pansharpening datasets used by DL methods. Resolutions and sizes refer to the target fusion image.}
\centering
\scriptsize
\setlength{\tabcolsep}{1mm}
\begin{tabular}{l@{\hspace{10mm}}l@{\hspace{15mm}}c@{\hspace{5mm}}c@{\hspace{5mm}}c@{\hspace{5mm}}c@{\hspace{5mm}}cc}
\hline
\multirow{2}{*}{Work} & \multirow{2}{*}{Instrument - Scene} & \multirow{2}{*}{HS}& \multirow{2}{*}{PAN}& \multirow{2}{*}{$R$} & \multirow{2}{*}{Resolution} & Train/Valid. Patches & Test Patches \\
& & & &                            &                         & $N\times \left(H \times W\right)$      & $N\times \left(H \times W\right)$ \\
\hline\hline
\multirow{3}{*}{HyperPNN \cite{He2019a}} & HYDICE - Washington DC Mall & $\H_\downarrow$ & $\langle\H\rangle$ &5  &2 m &n.a.$\times$(11$\times$11)$^{\rm (o)}$& 1$\times$256$\times$128 \\
 & AVIRIS - Moffett Field & $\H_\downarrow$ & $\langle\H\rangle$&5 &20 m& n.a.$\times$(11$\times$11$)^{\rm (o)}$& 1$\times$(256$\times$128)\\
& AVIRIS - Salinas Valley& $\H_\downarrow$ & $\langle\H\rangle$&5 &3.7 m& n.a.$\times$(11$\times$11)$^{\rm (o)}$& 1$\times$(256$\times$128)\\
\hline
\multirow{5}{*}{HSpeNet \cite{He2020}} & HYDICE - Washington DC Mall& $\H_\downarrow$ &$\langle\H\rangle$ & 5 & 2 m& 4284$\times$(11$\times$11)$^{\rm (o)}$& 1$\times$(200$\times$200) \\
 & AVIRIS - Moffett Field& $\H_\downarrow$ &$\langle\H\rangle$& 5 &20 m& 1645$\times$(11$\times$11)$^{\rm (o)}$& 1$\times$(150$\times$150)\\
 & ROSIS - Pavia University& $\H_\downarrow$ & $\langle\H\rangle$& 5&1.3 m&3120$\times$(11$\times$11)$^{\rm (o)}$& 1$\times$(200$\times$200)\\
 & EO-1/ALI - Halls Creek & $\H_\downarrow$ & $\P_\downarrow$& 3 &30 m &n.a.$\times$(11$\times$11)$^{\rm (o)}$& 1$\times$(120$\times$120)\\
 & EO-1/ALI - Halls Creek& real  & real& 3 & 10 m& n.a.$\times$(11$\times$11)$^{\rm (o)}$ & n.a.\\
\hline
\multirow{4}{*}{DHP-DARN \cite{Zheng2020}} & CAVE (real world scenes)& $\H_\downarrow$  &$\langle\H\rangle$& 4 &n.a.&5632$\times$(32$\times$32)& 10$\times$(512$\times$512) \\
 & ROSIS - Pavia Center& $\H_\downarrow$  & $\langle\H\rangle$& 4& 1.3 m &425$\times$(32$\times$32)& 7$\times$(160$\times$160)\\
 & EO-1 - Botswana& $\H_\downarrow$  &$\langle\H\rangle$& 3 & 30 m & 126$\times$(32$\times$32)& 6$\times$(120$\times$120)\\
 & EO-1/ALI - Los Angeles& real  & real & 3 &10 m & &1$\times$360$\times$360\\
\hline
\multirow{4}{*}{DIP-HyperKite \cite{Bandara2022}} & ROSIS - Pavia Center& $\H_\downarrow$ & $\langle\H\rangle$& 4& 1.3 m& 17$\times$(160$\times$160)& 7$\times$(160$\times$160)\\
&MV.C VNIR - Chikusei & $\H_\downarrow$ &$\langle\H\rangle$& 4& 2.5 m& 61$\times$(256$\times$256)& 20$\times$(256$\times$256)\\
&EO-1 - Botswana& $\H_\downarrow$ &$\langle\H\rangle$& 3 & 30 m& 14$\times$(120$\times$120)& 6$\times$(120$\times$120)\\
&EO-1/ALI - Los Angeles & real & real & 3& 10 m & &1$\times$(360$\times$360)\\
%\hline
%\multirow{2}{*}{RE-RANet \cite{Nie2022}}
%&EO-1/ALI - n.a. & $\H_\downarrow$ &  real &162/242&30m& n.a ${\times}$ n.a  & 3 & &\red{93\%$^\dag$}&&\red{7\%$^\dag$}\\
%&MV.C VNIR - Chikusei & $\H_\downarrow$ &$\langle\H\rangle$& 124/512 & 2.5m & 2517 $\times$ 2334  & 3 & 839 $\times$ 778 & \red{3\%} & & \red{1\%}\\
\hline
\multirow{5}{*}{HyperDSNet \cite{Zhuo2022}} & HYDICE - Washington DC Mall & $\H_\downarrow$ &$\langle\H\rangle$& 4& 2 m& 1024$\times$(64$\times$64)$^{\rm (o)}$& 4$\times$(128$\times$128) \\
 & ROSIS - Pavia Center& $\H_\downarrow$  &$\langle\H\rangle$& 4& 1.3 m & 1680$\times$(64$\times$64)$^{\rm (o)}$& 2$\times$(400$\times$400)\\
 & EO-1 - Botswana& $\H_\downarrow$  &$\langle\H\rangle$& 4 & 30 m &  967$\times$(64$\times$64)$^{\rm (o)}$& 4$\times$(128$\times$128)\\
 & PRISMA - Bologna& $\H_\downarrow$  & $\P_\downarrow$& 6 & 30 m & 816$\times$(60$\times$60)$^{\rm (o)}$&  \\
 & PRISMA - Bologna& real  & real & 6 & 5 m &  & 2$\times$(240$\times$240) \\
\hline
\multirow{5}{*}{R-PNN \cite{Guarino2023}}   & PRISMA - Barcelona & $\H_\downarrow$ & $\P_\downarrow$& 6 & 30 m &  &1$\times$(900$\times$900)\\
&PRISMA - Milan & $\H_\downarrow$ &  $\P_\downarrow$& 6 & 30 m & &1$\times$(900$\times$900)\\
&PRISMA - Prato & real  &real& 6& 5 m & 100$\times$(240$\times$240)$^{\rm (f)}$&\\
&PRISMA - Bologna & real  &real& 6 & 5 m & &1$\times$(2400$\times$2400)\\
&PRISMA - Florence & real  &real& 6 & 5 m& &1$\times$(2400$\times$2400)\\
 \hline
\multirow{6}{*}{PCA-Z-PNN \cite{Guarino2023a}}   &PRISMA - Milan & $\H_\downarrow$ &$\P_\downarrow$& 6& 30 m & 1$\times$(900$\times$900)$^{\rm (v)}$&\\
&PRISMA - Barcelona& $\H_\downarrow$ &$\P_\downarrow$& 6& 30 m& &1$\times$(900$\times$900)\\
&ROSIS - Pavia University & $\H_\downarrow$ &$\langle\H\rangle$& 6 & 1.3 m& &1$\times$(606$\times$336)\\
&PRISMA - Prato & real &real& 6& 5 m&  1$\times$(2400$\times$2400$)^{\rm (v)}$&\\
&PRISMA - Bologna & real & real& 6 &5 m& &1$\times$(2400$\times$2400)\\
&PRISMA - Florence & real &real& 6&5 m &  &1$\times$(2400$\times$2400)\\
\hline
\multicolumn{8}{l}{\rule{0pt}{8pt}$^{\rm (v)}$: Validation only \hs $^{\rm (f)}$: First band only
\hs $^{\rm (o)}$: With overlap \hs $\langle\cdot\rangle$: Bands average \hs $\downarrow$: Resolution downgrade
\hs n.a.: Info not available}
\end{tabular}
\label{tab:datasets}
\end{table*}

Before presenting the proposed benchmarking dataset, it is worth reviewing the most popular datasets used for HS pansharpening evaluation.
Since this topic is particularly important for DL-based methods, we limit attention to the latter, and in particular to those implemented for the toolbox.
In any case, these datasets are quite representative of the overall literature (DL or not) on HS pansharpening.
%They are gathered in Tab.~\ref{tab:datasets}, with all the details on their characteristics and use.
Tab.~\ref{tab:extra_det} summarizes the main features of such datasets (bands, spectral range, size, documentation, availability), 
while Tab.~\ref{tab:datasets} provides
useful information about their use by different research teams.
For each work/dataset, the HS and PAN columns indicate whether these two components are real (as distributed by the provider) or synthetic.
The latter are obtained through spatial resolution downgrading ($\H_\downarrow$, $\P_\downarrow$) or by averaging the HS bands, limiting the scope to the visible spectrum ($\langle\H\rangle$).
In most cases, the use of synthetic components is unavoidable because the original image is not multi-resolution and only the HS component is given, with no PAN.
Sometimes, instead, truly multiresolution data are available, like the EO-1/ALI combination or the PRISMA images.
In these cases, two options are possible:
{\it  i)} downgrade the resolution of both HS and PAN anyway, in order to use supervised training and full-reference quality metrics;
{\it ii)} keep the original data to preserve information, and use some forms of unsupervised training together with no-reference quality metrics.

Continuing to scan the columns of Tab.~\ref{tab:datasets},
we first find the resolution ratio, $R$, which goes from 3 to 6, the most challenging case. Actually, this value is arbitrary for all datasets without a real PAN component.
Then, the number of bands (used/all) points out the critical issue of image structure variability that occurs not only for different sensors but also for different scenes acquired with the same instrument.
The spectral range covered is rather uniform, going from the visible (0.4 $\mu$m on) to the short-wave infrared (up to 2.4-2.5 $\mu$m), with the exception of ROSIS and MV.C VNIR, which cover up to the near-infrared and CAVE, limited to the visible spectrum.
The physical size of the pixels, known as spatial resolution or ground sample distance (GSD), is also quite variable, ranging from 1.3~m up to 30~m, referring to the pansharpened image domain, and accounting also for resolution downgrade if any.
Sometimes, the resolution is very high.
This is due to the use of airborne instruments ({\it e.g.}, AVIRIS) operated at a much lower altitude than satellite ones.
This is a further source of variability that must be taken into account.

The last three columns concern the dataset size, measured again in the output domain and including possible resolution downgrade.
After the image size, its partitioning in training\footnote{More precisely training {\em and} validation, just training for brevity.} and test subsets is specified, together with its organization in patches, all relevant pieces of information for DL-based methods.
The first five methods, from HyperPNN to HyperDSNet, based on supervised learning,
experiment almost exclusively on RR images, using (non-overlapping) fragments of the same image for both training and testing.
This practice, due to the inability to manage a number of bands that varies from image to image, makes it hard to assess the generalization ability of the methods.
However, almost all supervised approaches present at least a test experiment on FR real images.
In \cite{He2020} the model HSpeNet, trained on the RR version of the EO-1/ALI - Halls Creek dataset, is then tested on the FR version with an assessment limited to the visual inspection.
A similar test, using the RR and FR versions of the PRISMA - Bologna dataset, is proposed in \cite{Zhuo2022} for the HyperDSNet model.
In this case, a numerical assessment of the spatial and spectral consistency is given.
In \cite{Zheng2020} and \cite{Bandara2022}, the networks DHP-DARN and DIP-HyperKite, trained on the RR Botswana image, were tested on the FR Los Angeles image, assessing numerically both the spatial and spectral consistencies.
The last two methods work in the target-adaptive modality, optimizing their parameters, without supervision, directly on the test image.
R-PNN only needs a single band (lowest wavelength) for pre-training, while PCA-Z-PNN requires none, as the adaptation starts from random weights.
Therefore, different images can be used for training and testing,
and the assessment is carried out both at reduced resolution, with full-reference metrics and at the original resolution, based on the spectral and spatial distortion indices.

The analysis of Tab.~\ref{tab:datasets} shows a rather puzzling scenario.
Different groups use different test images, so the results are not immediately comparable with the prior art.
Furthermore, they train models on images characterized by a number of bands that changes from case to case, so most of the trained models cannot be used on new images with a different number of bands.
This, in turn, prevents the generalization ability of the methods to be verified.
Finally, training sets are often just too small to ensure correct training.

Some problems cannot be solved with the use of a better dataset, just because sensors are often incompatible with one another.
However, limiting attention to the case of a fixed instrument, an ideal dataset for HS pansharpening should at least satisfy the following criteria:
\begin{itemize}
\item   it should include many acquisitions, with as much diversity as possible in terms of geographical regions, land covers, acquisition conditions;
\item   training and test sets should not share the same acquisitions;
\item   both the training (especially) and the test set should include much more than a single image;
\item   all images should share the same set of spectral bands for cross-image validation and testing and to allow fair comparison between all methods, regardless of their ability to handle a variable number of bands;
\item    images should be truly multi-resolution, with a high-resolution PAN associated with the HS image, to enable real-world full-resolution testing;
\item   the dataset should be freely available to the community to ensure its widest use and reproducibility of results.
\end{itemize}

\subsection{The proposed dataset}
A major contribution of this work is to provide a dataset that meets all the above requirements.
Considering the restrictive data-sharing policies widespread in the remote sensing field, we decided to use the PRISMA (PRecursore IperSpettrale della Missione Applicativa) images,
shared on-demand by the Italian Space Agency (ASI) for research purposes only.
In particular, a set of 16 images has been carefully selected and organized as summarized in Tab.~\ref{tab:prisma},
including both the full-resolution and reduced-resolution versions.
The first set of 12 images is reserved for training and validation purposes (part A of the table).
All images are rather large, 3600$\times$3600 pixels at the target 5m resolution, which allows us to extract 10 or more large tiles from each of them,
with size 1152$\times$1152 at full resolution and 192$\times$192 after the 6$\times$ decimation needed for RR training.
As always happens with HS images, not all bands have sufficient quality for further processing.
Therefore, out of the 239 bands available, we identified a subset of 159 bands that have good quality in all training and test images.
A second set of 4 images, is kept for testing (part B) and form the strict-sense benchmarking dataset.
Again, both full and reduced-resolution versions are given.
In the former case, only 1200$\times$1200 crops are used, to limit complexity, while the complete 600$\times$600 RR images are used in the latter case.
The images have been acquired over a period of more than 3 years, all over the world, and display a wide variety of land covers.
In Fig.~\ref{fig:prisma} we show an RGB composition of the original HS images. 
For RR tests these are used in their full size, whereas smaller tiles (red solid line-boxes) are considered for FR tests to limit the computation time.
In fact, in the FR space each HS test image would get a size of 3600$\times$3600$\times$159, once interpolated to match the target spatial size,
giving rise to a considerable volume.
The tiles are instead 200$\times$200 pixels wide,
hence 1200$\times$1200 with interpolation, {\em i.e.} 4 times larger images compared to the RR test images. Red and green dashed line-boxes highlight
the crops sampled for visualization purposes in the experimental section
for FR and RR tests, respectively.

%\begin{figure*}
%    \centering
%    \setlength\tabcolsep{1pt}{
%    \begin{tabular}{cccc}
%    \includegraphics[width=0.24\linewidth]{./prisma_img/20220905101901RR_GT_RGB.jpeg}
%    \includegraphics[width=0.24\linewidth]{./prisma_img/20230824100356RR_GT_RGB.jpeg}
%    \includegraphics[width=0.24\linewidth]{./prisma_img/20230908173127RR_GT_RGB.jpeg}
%    \includegraphics[width=0.24\linewidth]{./prisma_img/20231120102229RR_GT_RGB.jpeg}
%    \end{tabular}
%    }
%    \caption{Caption.}
%    \label{label}
%\end{figure*}

\newcommand{\wid}{0.35}
\begin{figure*}
    \centering
    \setlength\tabcolsep{6pt}
    \begin{tabular}{ccc}
    \includegraphics[width=\wid\linewidth]{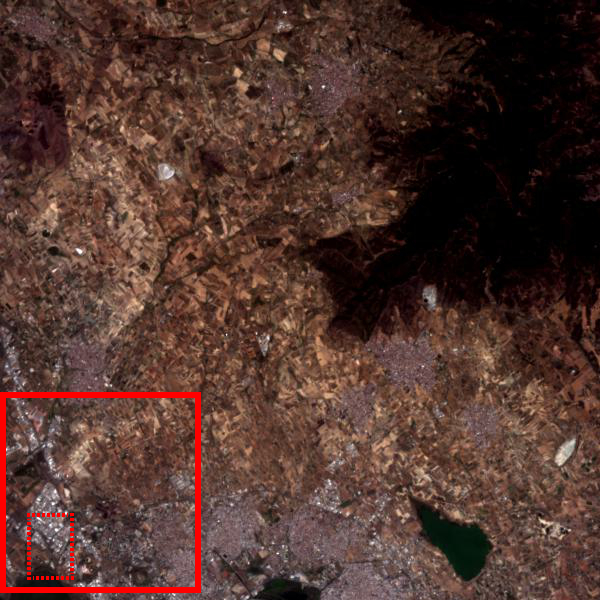} &
    \includegraphics[width=\wid\linewidth]{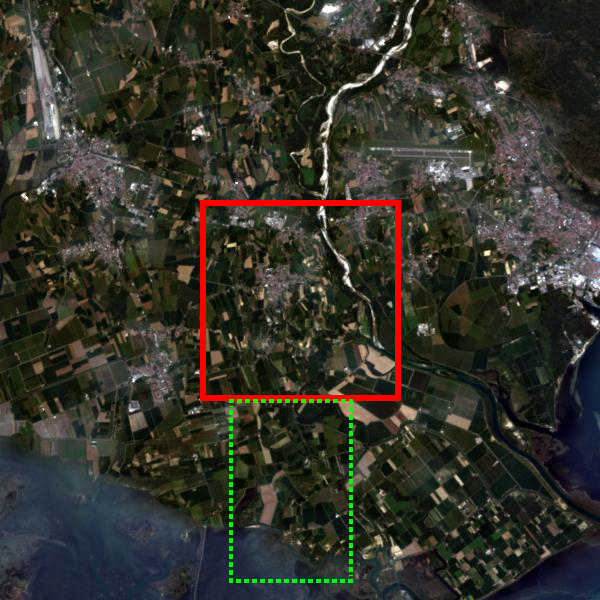} \\
    Cagliari, Italy & Udine, Italy \\[1mm]
    \includegraphics[width=\wid\linewidth]{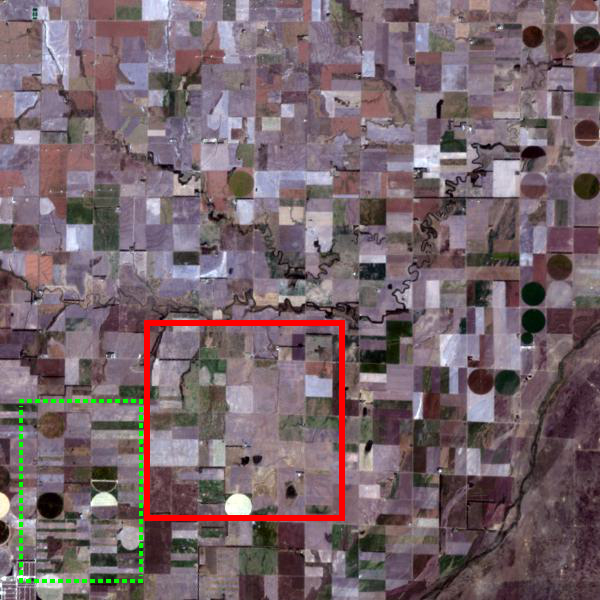} &
    \includegraphics[width=\wid\linewidth]{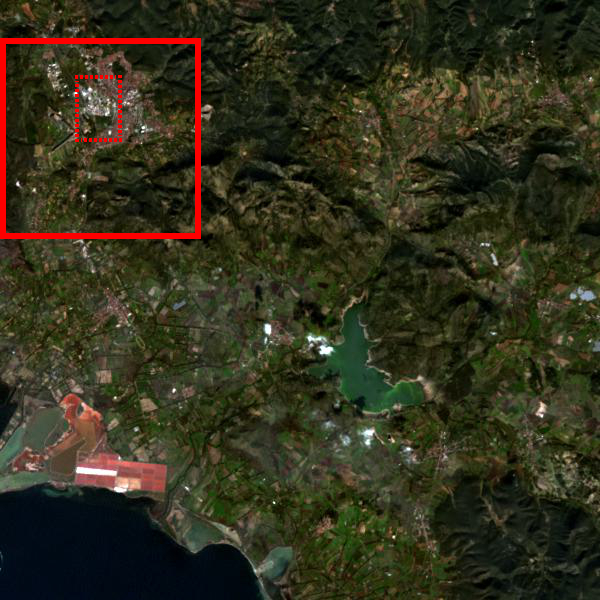}\\
    Ford County, Kansas, USA & Macuspana, Tabasco, MEX
    \end{tabular}
    \caption{Test set. HS (RGB bands) PRISMA test images.
    Red solid line-boxes indicate the tiles for FR tests. 
    RR tests are carried out on the whole images.
    Dashed line-boxes spot the crops used for visualization purposes in the experimental part for both RR (green) and FR (red) tests.}
    \label{fig:prisma}
\end{figure*}

\begin{table}
\caption{PRISMA benchmark datasets: (A) Training, (B) Test.}
\centering
\scriptsize
\setlength{\tabcolsep}{1mm}
\begin{tabular}{c}
\begin{tabular}{llcl}
%\hline
%\multicolumn{4}{c}{TRAIN+VALIDATION}\\
\hline
%&&Train&\\
Date     & Location               & Patches & Land covers\\ \hline
27/06/20 & Florence, Italy        & 14 & Roofs, Streets, Crops, Hills, Water\\
13/08/20 & Milan, Italy           & 13 & Roofs, Streets, Crops\\
07/11/20 & Bologna, Italy         & 13 & Roofs, Streets, Crops, Hills\\
03/03/21 & Kanpur, India          & 14 & Roofs, Streets, Crops, Hills, Rivers\\
03/07/21 & Willow Hill, Illinois  & 14 &        Streets, Crops, Hills, Rivers\\
19/06/22 & Onaka, South Dakota    & 14 &                 Crops,        Swamps\\
03/07/22 & Grosseto, Italy        & 14 & Roofs, Streets, Crops, Hills\\
04/08/22 & Bordeaux, France       & 14 & Roofs, Streets, Crops\\
01/06/23 & Dijon, France          & 13 & Roofs, Streets, Crops\\
05/09/23 & Ontario, Canada        & 14 & Roofs, Streets, Crops, Hills, Rivers\\
09/09/23 & London, UK             & 10 & Roofs, Streets, Crops,        Water\\
20/11/23 & South Sardinia, Italy  & 16 & Roofs, Streets, Crops, Hills, Water \\
\hline
\multicolumn{2}{l}{Total \# of training patches} & 163& \\
\hline
\multicolumn{2}{l}{Total \# of validation patches} & 24& (2/zone)\\
\hline
\multicolumn{4}{l}{Patch size for supervised training (RR): 192$\times$192}\\
\multicolumn{4}{l}{Patch size for unsupervised training (FR): 1152$\times$1152}\\
%\hline
\multicolumn{4}{l}{Resolution ratio $R=6$}\\
%\hline
\multicolumn{4}{l}{Bands $B=159/239$}\\
%\hline
\multicolumn{4}{l}{Resolution: 30 m (RR) or 5 m (FR)}\\
\hline
\end{tabular}
\\
\rule{0mm}{8pt}(A)
\\
\\
\begin{tabular}{lll}
%\hline
\hline
Date & Location & Land covers\\
\hline
05/09/22 & Cagliari, Italy &  Roofs, Streets, Crops, Hills, Water\\
24/08/23 & Udine, Italy &  Roofs, Streets, Crops, Water\\
08/09/23 & Ford County,  Kansas & Streets, Crops\\
20/11/23 & Macuspana, Tabasco & Streets, Crops \\
\hline
\multicolumn{3}{l}{RR test image size: 600$\times$600 (from a 3600$\times$3600 real FR image)}\\
\multicolumn{3}{l}{Real FR test image size: 1200$\times$1200}\\
\multicolumn{3}{l}{Resolution ratio $R=6$}\\
\multicolumn{3}{l}{Bands $B=159/239$}\\
\multicolumn{3}{l}{Resolution: 30 m (RR) or 5 m (FR)}\\
\hline

\end{tabular}
\\
\rule{0mm}{8pt}(B)
\end{tabular}
\label{tab:prisma}
\end{table}

\subsection{Is this worth the effort?}
We have already underlined what possible advantages HS pansharpening research can benefit from the use of a large and well-structured shared dataset.
Here, we perform a very simple yet enlightening experiment in RR space to support our claims.

As noted above,
in the absence of suitable datasets, a common practice for DL-based methods is to use a single image split into two non-overlapping parts for training and testing.
To simulate this condition, we train our 5 supervised models, following the original experimental configurations,
on a fraction (about 3/4) of the Udine image, listed in Tab.\ref{tab:prisma}(B), and test them on the remaining part, call these parts Udine/1 and Udine/2.
Numerical results are reported in Tab.~\ref{tab:ISExp} (only ERGAS for brevity).
All five methods work quite well (on visual inspection) with very close ERGAS indicators, ranging from 1.7 to 2.1.
Then, we train the same models on the proposed training set, see Tab.~\ref{tab:prisma}(A),
obtaining slightly worse results in all cases\footnote{Much worse for DHP-DARN, which seems to be an outlier in this case, {\em i.e.}, 
extremely sensitive to the training set.}.
As expected the statistical ``alignment'' between training and testing data offers some advantages.
In fact, even if Udine/1 and Udine/2 are disjoint,
they come from the same acquisition and share the same sensing geometry, daylight and atmosphere conditions, land cover, etc.

Things change completely when models are tested on a new image, Cagliari, again in Tab.\ref{tab:prisma}(B).
The models trained on the proposed dataset exhibit a rather stable performance, with a slight average improvement concerning Udine/2.
Neither Udine nor Cagliari contribute to the training set, so these results suggest that the latter can more easily pansharpened because of its simpler structure (see also Fig.~\ref{fig:prisma}).
However, the models trained on Udine/1 suffer a catastrophic impairment, with ERGAS increasing by more than 80\% on average.
This is because their weights, over-fit to a small training set, are unable to generalize well to new data.
In summary, this experiment, although limited in scope,
confirms that the use of large, well-structured datasets improves generalization and prevents biases in experimental results.

\newcommand{\mr}[1]{\multirow{2}{*}{#1}}
\begin{table}
\centering
\caption{ERGAS scores with small/aligned {\em vs} large training datasets.}

\begin{tabular}{lccc} \hline
\ru                &                  &  \multicolumn{2}{c}{Test image} \\ \cline{3-4}
method             & Training         &      Udine/2   & Cagliari   \ru \\ \hline
\mr{HyperPNN}      & Udine/1          &     \bf 1.8293 &     2.9550 \ru \\
                   & proposed dataset &         2.1704 & \bf 1.5802 \ru \\ \hline
\mr{HSpeNet}       & Udine/1          &     \bf 1.6823 &     3.6121 \ru \\
                   & proposed dataset &         1.8948 & \bf 1.2789 \ru \\ \hline
\mr{DIP-HyperKite} & Udine/1          &     \bf 1.8905 &     3.3532 \ru \\
                   & proposed dataset &         2.0212 & \bf 2.4194 \ru \\ \hline
\mr{DHP-DARN}      & Udine/1          &     \bf 2.0758 &     4.6525 \ru \\
                   & proposed dataset &         4.9047 & \bf 2.0733 \ru \\ \hline
\mr{Hyper-DSNet}   & Udine/1          &     \bf 1.7190 &     2.5472 \ru \\
                   & proposed dataset &         1.7748 & \bf 1.2298 \ru \\ \hline
\end{tabular}

\label{tab:ISExp}
\end{table}

\begin{table*}

%\caption{Reduced resolution results. \bestavg{Best}, \bestfiveavg{Top-5} and \worstfiveavg{Worst-5} are in \bestavg{bold green}, \bestfiveavg{green} and \worstfiveavg{red}, respectively.
%For each index the average score over the four test datasets is reported in the rightmost column.}
\caption{Results at Reduced Resolution. \bestavg{Best}, \bestfiveavg{Top-5} and \worstfiveavg{Worst-5} are in \bestavg{bold green}, \bestfiveavg{green} and \worstfiveavg{red}. The average score is in the rightmost column.}

\footnotesize
\centering
\setlength{\tabcolsep}{2pt}
\begin{tabular}{lc@{\rule{2mm}{0mm}}cccccc@{\rule{2mm}{0mm}}cccccc@{\rule{2mm}{0mm}}cccccc} \hline
\ru Method	&	&		\multicolumn{5}{c}{ERGAS}																		&	&		\multicolumn{5}{c}{SAM}																		&	&		\multicolumn{5}{c}{Q$2^n$}																		 \\	\hline
%\ru 	&	&		$\#1$		&		$\#2$		&		$\#3$		&		$\#4$		&		Avg.		&	&		$\#1$		&		$\#2$		&		$\#3$		&		$\#4$		&		Avg.		&	&		$\#1$		&		$\#2$		&		$\#3$		&		$\#4$		&		Avg.		 \\	\cline{1-1} \cline{3-7} \cline{9-13} \cline{15-19}
\ru	&	&		Cagliari		&		Udine		&		Ford		&		Tabasco		&		Avg.		&	&		Cagliari		&		Udine		&		Ford		&		Tabasco		&		Avg.		&	&		Cagliari		&		Udine		&		Ford		&		Tabasco		&		Avg.		 \\	\cline{1-1} \cline{3-7} \cline{9-13} \cline{15-19}
\ru (Ideal)	&	&		0		&		0		&		0		&		0		&		0		&	&		0		&		0		&		0		&		0		&		0		&	&		1		&		1		&		1		&		1		&		1		 \\	\cline{1-1} \cline{3-7} \cline{9-13} \cline{15-19}
\ru EXP	&	&		1.7716		&		3.9587		&		1.6921		&		4.3650		&		2.9468		&	&		2.3073		&		4.6288		&		2.8528		&		6.3524		&		4.0353		&	&		0.5971		&		0.6453		&		0.7984		&		0.5935		&	0.6586		 \\	\cline{1-1} \cline{3-7} \cline{9-13} \cline{15-19}
\ru GSA	&	&	\bestfiveim{0.9389}	&		2.4243		&	\bestfiveim{1.0008}	&		2.8065		&		1.7926		&	&	\bestfiveim{1.8191}	&		3.5730		&	\bestfiveim{2.2570}	&	\bestfiveim{5.7455}	&	\bestfiveavg{3.3486}	&	&		0.8739		&		0.8110		&		0.9201		&		0.7955		&		0.8501		 \\	
BT-H	&	&	\worstfiveim{1.2784}	&		3.4898		&		1.2462		&		3.1015		&		2.2790		&	&		2.3857		&		4.6059		&		2.5563		&		6.3532		&		3.9753		&	&	\worstfiveim{0.8180}	&		0.7166		&		0.8941		&		0.7837		&		\worstfiveavg{0.8031}		 \\	
BDSD-PC	&	&		1.1664		&		2.7627		&		1.2637		&	\worstfiveim{3.9316}	&		2.2811		&	&		1.9439		&		5.0116		&		2.6646		&	\worstfiveim{7.8908}	&		4.3777		&	&		0.8524		&		0.7904		&		0.9048		&	\worstfiveim{0.7144}	&		0.8155		 \\	
PRACS	&	&		1.1803		&		3.4688		&		1.2773		&		3.8578		&		2.4461		&	&		1.9742		&		4.2221		&		2.4571		&		6.2161		&		3.7174		&	&	\worstfiveim{0.8166}	&	\worstfiveim{0.7144}	&	\worstfiveim{0.8804}	&	\worstfiveim{0.6778}	&	\worstfiveavg{0.7723}	 \\	\cline{1-1} \cline{3-7} \cline{9-13} \cline{15-19}
\ru MTF-GLP-FS	&	&	\bestfiveim{0.9239}	&		2.3660		&	\bestfiveim{0.9808}	&	\bestfiveim{2.7356}	&	\bestfiveavg{1.7516}	&	&	\bestfiveim{1.8123}	&	\bestfiveim{3.4496}	&	\bestfiveim{2.2297}	&	\bestfiveim{5.7208}	&	\bestavg{3.3031}	&	&	\bestfiveim{0.8792}	&		0.8218		&	\bestfiveim{0.9238}	&		0.8026		&	\bestfiveavg{0.8568}	 \\	
MTF-GLP-HPM	&	&	\bestfiveim{0.9292}	&		4.0148		&		1.2231		&		3.0609		&		2.3070		&	&	\bestfiveim{1.8555}	&	\worstfiveim{5.1824}	&		2.6120		&		6.4104		&		4.0151		&	&	\bestfiveim{0.8833}	&	\worstfiveim{0.6698}	&		0.9000		&		0.7890		&		0.8105		 \\	
MTF-GLP-HPM-R	&	&	\bestim{0.9114}	&		2.4661		&	\bestim{0.9743}	&	\bestfiveim{2.6967}	&	\bestfiveavg{1.7621}	&	&	\bestim{1.8121}	&	\bestfiveim{3.3953}	&	\bestfiveim{2.2124}	&		6.0986		&	\bestfiveavg{3.3796}	&	&	\bestfiveim{0.8800}	&		0.8134		&	\bestfiveim{0.9239}	&		0.8023		&		0.8549		 \\	
AWLP	&	&		1.1846		&	\worstfiveim{4.6926}	&		1.3424		&	\worstfiveim{3.8592}	&	\worstfiveavg{2.7697}	&	&		2.3323		&	\worstfiveim{7.9280}	&	\worstfiveim{2.9252}	&		7.7668		&	\worstfiveavg{5.2381}	&	&		0.8516		&	\worstfiveim{0.6290}	&		0.8832		&	\worstfiveim{0.7496}	&	\worstfiveavg{0.7783}	 \\	
MF	&	&		1.1392		&	\worstfiveim{4.0155}	&	\worstfiveim{1.5506}	&		3.3944		&	\worstfiveavg{2.5249}	&	&		1.9213		&		4.5587		&		2.7173		&		6.2918		&		3.8723		&	&		0.8638		&		0.7263		&	\worstfiveim{0.8804}	&		0.7807		&		0.8128		 \\	\cline{1-1} \cline{3-7} \cline{9-13} \cline{15-19}
\ru HySURE	&	&	\worstfiveim{1.4664}	&	\worstfiveim{4.3476}	&	\worstfiveim{2.0749}	&	\worstfiveim{5.1119}	&	\worstfiveavg{3.2502}	&	&	\worstfiveim{2.9264}	&	\worstfiveim{6.6950}	&	\worstfiveim{4.4153}	&	\worstfiveim{9.0468}	&	\worstfiveavg{5.7709}	&	&	\worstfiveim{0.7943}	&	\worstfiveim{0.5761}	&	\worstfiveim{0.7475}	&	\worstfiveim{0.4951}	&	\worstfiveavg{0.6532}	 \\	
SR-D	&	&	\worstfiveim{1.8476}	&	\worstfiveim{4.2618}	&	\worstfiveim{1.8825}	&	\worstfiveim{4.8011}	&	\worstfiveavg{3.1982}	&	&	\worstfiveim{2.4242}	&	\worstfiveim{6.1149}	&	\worstfiveim{3.0551}	&	\worstfiveim{7.9927}	&	\worstfiveavg{4.8967}	&	&	\worstfiveim{0.5734}	&	\worstfiveim{0.5846}	&	\worstfiveim{0.7313}	&	\worstfiveim{0.5440}	&	\worstfiveavg{0.6083}	 \\	
TV	&	&		1.2624	&		2.8864		&		1.3163		&		3.1771		&		2.1605		&	&		2.2212		&		4.0400		&		2.5705		&	\bestfiveim{5.6672}	&		3.6247		&	&	\worstfiveim{0.8291}	&		0.7833		&		0.8911		&		0.7692		&		0.8182		 \\	\cline{1-1} \cline{3-7} \cline{9-13} \cline{15-19}
\ru HyperPNN	&	&		1.2245		&	\bestfiveim{2.2237}	&		1.1864		&		3.3557		&		1.9976		&	&	\worstfiveim{2.8925}	&		4.3058		&		2.6461		&		7.0296		&		4.2185		&	&	\bestfiveim{0.8777}	&	\bestfiveim{0.8539}	&	\bestfiveim{0.9247}	&	\bestfiveim{0.8229}	&	\bestfiveavg{0.8698}	 \\	
HSpeNet	&	&		1.0354		&	\bestfiveim{2.0527}	&		1.0674		&		2.7964		&	\bestfiveavg{1.7379}	&	&		2.0644		&	\bestfiveim{3.2157}	&	\bestfiveim{2.2838}	&		5.8462		&	\bestfiveavg{3.3525}	&	&		0.8674		&	\bestfiveim{0.8588}	&		0.9136		&	\bestfiveim{0.8245}	&	\bestfiveavg{0.8660}	 \\	
DHP-DARN	&	&	\worstfiveim{1.9137}		&	\worstfiveim{4.4280}	&	\worstfiveim{1.7389}	&	\worstfiveim{4.7617}	&	\worstfiveavg{3.2106}	&	&	\worstfiveim{3.4673}	&		5.1740		&	\worstfiveim{3.3833}	&	\worstfiveim{9.1288}	&	\worstfiveavg{5.2883}	&	&		0.8489		&	\bestfiveim{0.8234}	&		0.9190		&		0.7934		&		0.8462		 \\	
DIP-HyperKite	&	&	\worstfiveim{1.3510}	&	\bestfiveim{2.0206}	&		1.0995		&		2.8092		&		1.8201		&	&	\worstfiveim{3.6543}	&	\worstfiveim{5.8268}	&		2.7089		&	\worstfiveim{9.1932}	&	\worstfiveavg{5.3458}	&	&		0.8419		&	\bestfiveim{0.8497}	&	\bestfiveim{0.9205}	&	\bestfiveim{0.8201}	&	\bestfiveavg{0.8581}	 \\	
Hyper-DSNet	&	&	\bestfiveim{0.9725}	&	\bestim{1.8198}	&	\bestfiveim{0.9820}	&	\bestim{2.4237}	&	\bestavg{1.5495}	&	&	\bestfiveim{1.9162}	&	\bestim{2.9638}	&	\bestim{2.1906}	&		7.4854		&		3.6390		&	&	\bestim{0.8833}	&	\bestim{0.8767}	&	\bestim{0.9303}	&	\bestim{0.8426}	&	\bestavg{0.8832}	 \\	
R-PNN	&	&		0.9884		&	\bestfiveim{2.2256}	&	\bestfiveim{1.0581}	&	\bestfiveim{2.6704}	&	\bestfiveavg{1.7356}	&	&		1.9720		&	\bestfiveim{3.5728}	&		2.3860		&	\bestfiveim{5.6797}	&	\bestfiveavg{3.4026}	&	&		0.8712		&		0.8225		&		0.9172		&		0.8112		&		0.8555		 \\	
PCA-Z-PNN	&	&		1.0824		&		2.9062		&	\worstfiveim{1.5575}	&	\bestfiveim{2.7813}	&		2.0819		&	&		2.1124		&		4.6125		&	\worstfiveim{3.1154}	&	\bestim{5.5297}	&		3.8425		&	&		0.8507		&		0.7798		&	\worstfiveim{0.8647}	&	\bestfiveim{0.8116}	&		0.8267		 \\	\hline
\end{tabular}

\label{tab:RR}
\end{table*}

\begin{table*}
%\caption{Full resolution results. \bestavg{Best}, \bestfiveavg{Top-5} and \worstfiveavg{Worst-5} are in \bestavg{bold green}, \bestfiveavg{green} and \worstfiveavg{red}, respectively.
%For each index the average score over the four test datasets is reported in the rightmost column.}
\caption{Results at Full Resolution. \bestavg{Best}, \bestfiveavg{Top-5} and \worstfiveavg{Worst-5} are in \bestavg{bold green}, \bestfiveavg{green} and \worstfiveavg{red}. The average score is in the rightmost column.}
\footnotesize
\centering
\setlength{\tabcolsep}{2pt}
\begin{tabular}{lc@{\rule{2mm}{0mm}}cccccc@{\rule{2mm}{0mm}}cccccc@{\rule{2mm}{0mm}}cccccc} \hline
\ru Method	&	&		\multicolumn{5}{c}{$\DL$}																		&	&		\multicolumn{5}{c}{$\DS$}																		&	&		\multicolumn{5}{c}{RQNR}																		 \\	\hline
\ru	&	&		Cagliari		&		Udine		&		Ford		&		Tabasco		&		Avg.		&	&		Cagliari		&		Udine		&		Ford		&		Tabasco		&		Avg.		&	&		Cagliari		&		Udine		&		Ford		&		Tabasco		&		Avg.		 \\	\cline{1-1} \cline{3-7} \cline{9-13} \cline{15-19}
\ru (Ideal)	&	&		0		&		0		&		0		&		0		&		0		&	&		0		&		0		&		0		&		0		&		0		&	&		1		&		1		&		1		&		1		&		1		 \\	\cline{1-1} \cline{3-7} \cline{9-13} \cline{15-19}
\ru EXP	&	&		0.0126		&		0.0077		&		0.0054		&		0.0106		&		0.0091		&	&		0.1306		&		0.1273		&		0.0704		&		0.1947		&		0.1308		&	&		0.8585		&		0.8660		&		0.9245		&		0.7967		&		0.8614		 \\	\cline{1-1} \cline{3-7} \cline{9-13} \cline{15-19}
\ru GSA	&	&		0.0227		&		0.0134		&		0.0078		&		0.0122		&		0.0140		&	&		0.0056		&		0.0074		&		0.0077		&		0.0184		&		0.0098		&	&	\bestfiveim{0.9718}	&	\bestfiveim{0.9793}	&	\bestfiveim{0.9846}	&	\bestfiveim{0.9697}	&	\bestfiveavg{0.9763}	 \\	
BT-H	&	&	\worstfiveim{0.0633}	&	\worstfiveim{0.0936}	&		0.0195		&	\worstfiveim{0.0288}	&	\worstfiveavg{0.0513}	&	&	\bestim{0.0000}	&	\bestim{0.0000}	&	\bestim{0.0000}	&	\bestim{0.0002}	&	\bestavg{0.0001}	&	&	\worstfiveim{0.9367}	&	\worstfiveim{0.9064}	&		0.9805		&	\bestfiveim{0.9710}	&		0.9487		 \\	
BDSD-PC	&	&	\worstfiveim{0.0439}	&	\worstfiveim{0.0256}	&		0.0220		&	\worstfiveim{0.0487}	&	\worstfiveavg{0.0351}	&	&	\bestfiveim{0.0024}	&	\bestfiveim{0.0000}	&	\bestfiveim{0.0001}	&	\bestfiveim{0.0003}	&	\bestfiveavg{0.0007}	&	&		0.9539		&		0.9744		&		0.9778		&		0.9510		&		0.9643		 \\	
PRACS	&	&	\bestfiveim{0.0118}	&		0.0155		&	\bestfiveim{0.0047}	&		0.0088		&		0.0102		&	&		0.0083		&		0.0163		&		0.0114		&		0.0245		&		0.0151		&	&	\bestim{0.9801}	&		0.9684		&	\bestfiveim{0.9840}	&	\bestfiveim{0.9670}	&	\bestfiveavg{0.9749}	 \\	\cline{1-1} \cline{3-7} \cline{9-13} \cline{15-19}
\ru MTF-GLP-FS	&	&	\bestfiveim{0.0101}	&	\bestfiveim{0.0054}	&	\bestfiveim{0.0033}	&	\bestfiveim{0.0065}	&	\bestfiveavg{0.0063}	&	&		0.0214		&		0.0277		&		0.0174		&		0.0322		&		0.0247		&	&		0.9686		&		0.9671		&		0.9794		&		0.9616		&		0.9692		 \\	
MTF-GLP-HPM	&	&		0.0147		&		0.0087		&		0.0067		&		0.0093		&		0.0098		&	&		0.0215		&	\worstfiveim{0.0284}	&		0.0179		&	\worstfiveim{0.0389}	&	\worstfiveavg{0.0267}	&	&		0.9641		&		0.9631		&		0.9756		&		0.9522		&		0.9637		 \\	
MTF-GLP-HPM-R	&	&	\bestfiveim{0.0100}	&	\bestfiveim{0.0057}	&	\bestfiveim{0.0033}	&	\bestfiveim{0.0074}	&	\bestfiveavg{0.0066}	&	&	\worstfiveim{0.0217}	&	\worstfiveim{0.0285}	&		0.0176		&		0.0374		&		0.0263		&	&		0.9685		&		0.9660		&		0.9791		&		0.9554		&		0.9673		 \\	
AWLP	&	&		0.0132		&		0.0097		&		0.0055		&		0.0092		&		0.0094		&	&	\worstfiveim{0.0251}	&	\worstfiveim{0.0318}	&	\worstfiveim{0.0206}	&	\worstfiveim{0.0407}	&	\worstfiveavg{0.0295}	&	&		0.9620		&		0.9588		&		0.9740		&		0.9504		&		0.9613		 \\	
MF	&	&	\worstfiveim{0.0655}	&	\worstfiveim{0.0392}	&	\worstfiveim{0.0390}	&	\worstfiveim{0.0331}	&	\worstfiveavg{0.0442}	&	&	\worstfiveim{0.0389}	&		0.0269		&	\worstfiveim{0.0250}	&	\worstfiveim{0.0664}	&	\worstfiveavg{0.0393}	&	&	\worstfiveim{0.8981}	&	\worstfiveim{0.9349}	&	\worstfiveim{0.9369}	&	\worstfiveim{0.9027}	&	\worstfiveavg{0.9182}	 \\	\cline{1-1} \cline{3-7} \cline{9-13} \cline{15-19}
\ru HySURE	&	&	\worstfiveim{0.1126}	&	\worstfiveim{0.0618}	&	\worstfiveim{0.0510}	&	\worstfiveim{0.1158}	&	\worstfiveavg{0.0853}	&	&	\bestfiveim{0.0022}	&	\bestfiveim{0.0011}	&	\bestfiveim{0.0016}	&	\bestfiveim{0.0022}	&	\bestfiveavg{0.0018}	&	&	\worstfiveim{0.8854}	&	\worstfiveim{0.9372}	&	\worstfiveim{0.9474}	&	\worstfiveim{0.8823}	&	\worstfiveavg{0.9131}	 \\	
SR-D	&	&		0.0128		&		0.0087		&		0.0055		&		0.0115		&		0.0096		&	&	\worstfiveim{0.1316}	&	\worstfiveim{0.1292}	&	\worstfiveim{0.0720}	&	\worstfiveim{0.1981}	&	\worstfiveavg{0.1327}	&	&	\worstfiveim{0.8573}	&	\worstfiveim{0.8632}	&	\worstfiveim{0.9229}	&	\worstfiveim{0.7927}	&	\worstfiveavg{0.8590}	 \\	
TV	&	&	\bestim{0.0042}	&	\bestim{0.0028}	&	\bestim{0.0021}	&	\bestim{0.0036}	&	\bestavg{0.0032}	&	&	\worstfiveim{0.0586}	&	\worstfiveim{0.0524}	&	\worstfiveim{0.0358}	&	\worstfiveim{0.0813}	&	\worstfiveavg{0.0570}	&	&		0.9375		&		0.9449		&	\worstfiveim{0.9622}	&	\worstfiveim{0.9154}	&	\worstfiveavg{0.9400}	 \\	\cline{1-1} \cline{3-7} \cline{9-13} \cline{15-19}
\ru HyperPNN	&	&	\worstfiveim{0.0415}		&		0.0139		&	\worstfiveim{0.0256}	&		0.0248		&		0.0264		&	&	\bestfiveim{0.0036}	&		0.0063		&	\bestfiveim{0.0040}	&	\bestfiveim{0.0067}	&	\bestfiveavg{0.0051}	&	&		0.9550		&	\bestfiveim{0.9799}	&		0.9705		&	\bestfiveim{0.9687}	&		0.9685		 \\	
HSpeNet	&	&		0.0174		&		0.0084		&	\worstfiveim{0.0240}	&		0.0154		&		0.0163		&	&		0.0101		&		0.0160		&		0.0141		&		0.0236		&		0.0159		&	&	\bestfiveim{0.9727}	&		0.9758		&		0.9623		&		0.9614		&		0.9680		 \\	
DHP-DARN	&	&	\worstfiveim{0.0832}	&	\worstfiveim{0.0615}	&	\worstfiveim{0.0467}	&	\worstfiveim{0.0513}	&	\worstfiveavg{0.0607}	&	&	\bestfiveim{0.0030}	&	\bestfiveim{0.0053}	&	\worstfiveim{0.0031}	&		0.0130		&		0.0061		&	&	\worstfiveim{0.9141}	&	\worstfiveim{0.9336}	&	\worstfiveim{0.9503}	&	\worstfiveim{0.9363}	&	\worstfiveavg{0.9336}	 \\	
DIP-HyperKite	&	&		0.0208		&		0.0125		&		0.0148		&		0.0146		&		0.0157		&	&		0.0042		&	\bestfiveim{0.0050}	&		0.0043		&	\bestfiveim{0.0059}	&	\bestfiveavg{0.0048}	&	&	\bestfiveim{0.9751}	&	\bestfiveim{0.9826}	&	\bestfiveim{0.9810}	&	\bestim{0.9796}	&	\bestavg{0.9796}	 \\	
Hyper-DSNet	&	&	\bestfiveim{0.0119}	&	\bestfiveim{0.0050}	&		0.0082		&	\bestfiveim{0.0068}	&	\bestfiveavg{0.0080}	&	&		0.0200		&		0.0246		&	\bestfiveim{0.0195}	&		0.0346		&		0.0247		&	&		0.9683		&		0.9705		&		0.9725		&		0.9588		&		0.9675		 \\	
R-PNN	&	&		0.0131		&		0.0091		&		0.0063		&	\bestfiveim{0.0074}	&		0.0090		&	&		0.0168		&		0.0144		&		0.0129		&		0.0341		&		0.0195		&	&		0.9704		&	\bestfiveim{0.9766}	&	\bestfiveim{0.9809}	&		0.9587		&	\bestfiveavg{0.9716}	 \\	
PCA-Z-PNN	&	&		0.0129		&	\bestfiveim{0.0041}	&	\bestfiveim{0.0046}	&		0.0080		&	\bestfiveavg{0.0074}	&	&		0.0152		&		0.0116		&		0.0088		&		0.0257		&		0.0153		&	&	\bestfiveim{0.9721}	&	\bestim{0.9844}	&	\bestim{0.9867}	&		0.9665		&	\bestfiveavg{0.9774}	 \\	\hline
\end{tabular}

\label{tab:FR}
\end{table*}

\section{Experimental results}
\label{sec:exp}
This work aims to provide a comprehensive overview of the current HS pansharpening framework.
This entails not only presenting and implementing classical and latest methods and evaluation metrics but also establishing guidelines for benchmarking and assessing these techniques, as well as highlighting open challenges and promising research directions in the field.
With these goals in mind, it is crucial to set the basic characteristics of an ideal HS pansharpening algorithm which should:
\begin{itemize}
\item[{(a)}] generalize across datasets;
\item[{(b)}] generalize across scales;
\item[{(c)}] preserve spectral features while raising spatial resolution;
\item[{(d)}] provide perceptually good visual results (for an ideal observer);
\item[{(e)}] be computationally efficient, especially at test time,
\end{itemize}
Properties (a)-(c) require reliable tools for numerical assessment, such as the SoTA indexes recalled in Section~\ref{sec:metrics},
while (d) remains a subjective evaluation, but is no less important.
Keeping the above properties in mind,
we aim to point out the strengths and especially the limitations of current SoTA algorithms, which represent the main open challenges for future research.
Note that we are not trying to establish some ranking among the techniques included in the benchmark, none of which is uniformly superior to the others.
Instead, we would like to suggest a performance analysis methodology as a guideline for future research in the field.

Let us now delve into the numerical results summarized in Tab.~\ref{tab:RR} and Tab.~\ref{tab:FR} for RR and FR datasets, respectively.
For easier understanding,
we have highlighted the 5 best and 5 worst performing techniques in green and red, respectively, with the best one emphasized in bold.
Comments to the results are gathered and ordered according to the above list of quality features.

\begin{figure*}
    \centering
    \setlength\tabcolsep{1pt}{
    \DD
    \begin{tabular}{cccccccccc}
        GT &   GSA &   BT-H &   BDSD-PC &   PRACS &   MTF-GLP-FS &   MTF-GLP-HPM &   MTF-GLP-HPM-R &   AWLP &   MF \\
        \imagerr{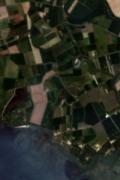} & \imagerr{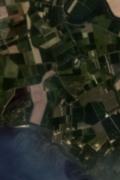} & \imagerr{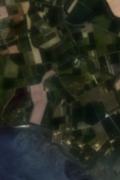} & \imagerr{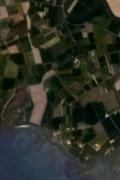} & \imagerr{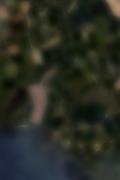} &
        \imagerr{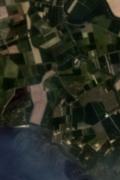} & \imagerr{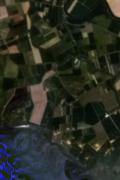} & \imagerr{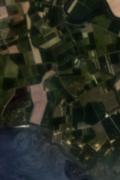} & 
         \imagerr{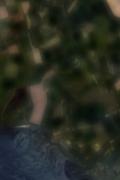} & \imagerr{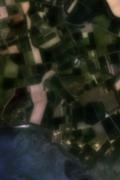} \\[1mm]
        HySURE &   SR-D &   TV &   HyperPNN &   HSpeNet &   DHP-DARN &   DIP-HyperKite &   Hyper-DSNet &   R-PNN &   PCA-Z-PNN \\
         \imagerr{20230824100356RR_HySure_RGB} & \imagerr{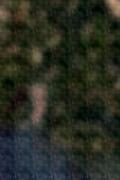} & \imagerr{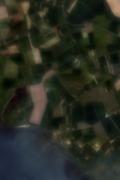} &
         \imagerr{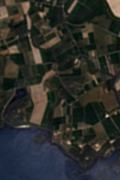} & \imagerr{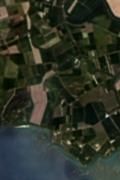} &
         \imagerr{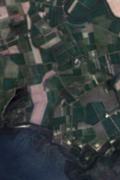} & \imagerr{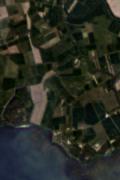} & \imagerr{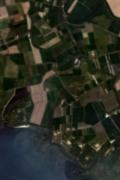} & \imagerr{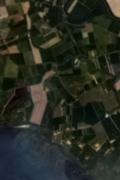} & \imagerr{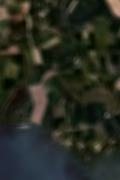} \\[1mm]
         \multicolumn{10}{c}{\D (a)} \\[1mm]
        GT &   GSA &   BT-H &   BDSD-PC &   PRACS &   MTF-GLP-FS &   MTF-GLP-HPM &   MTF-GLP-HPM-R &   AWLP &   MF \\
        \imagerr{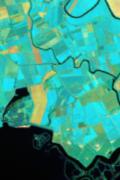} & \imagerr{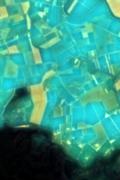} & \imagerr{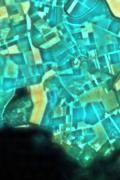} & \imagerr{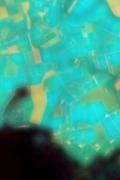} & \imagerr{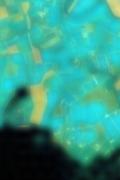} &
        \imagerr{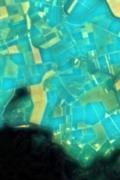} & \imagerr{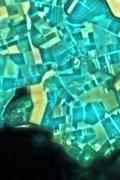} & \imagerr{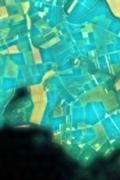} & 
         \imagerr{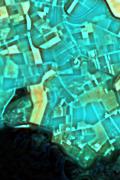} & \imagerr{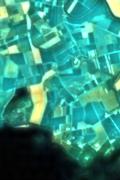} \\[1mm]
        HySURE &   SR-D &   TV &   HyperPNN &   HSpeNet &   DHP-DARN &   DIP-HyperKite &   Hyper-DSNet &   R-PNN &   PCA-Z-PNN \\
         \imagerr{20230824100356RR_HySure_FC} & \imagerr{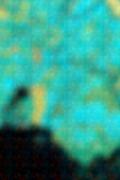} & \imagerr{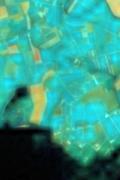} &
         \imagerr{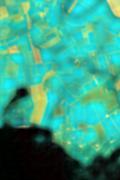} & \imagerr{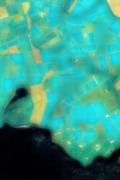} &
         \imagerr{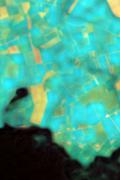} & \imagerr{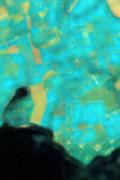} & \imagerr{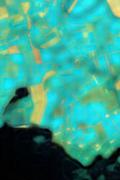} & \imagerr{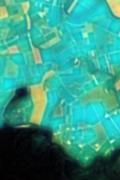} & \imagerr{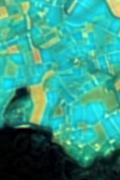} \\[1mm]
         \multicolumn{10}{c}{\D (b)} \\[1mm]
    \end{tabular}    
    }
    \caption{Pansharpening results (cropped: 120$\times$180) on Udine at reduced resolution. 
    Target GT (visible range on top, NIR-SWIR at bottom) 
    followed by all corresponding pansharpening results.
    Selected wavelenghts: $663, 560, 466$ nm (visible); $1943, 1261, 832$ nm (NIR-SWIR).}
    \label{fig:rr2}
\end{figure*}

\begin{figure*}
    \centering
    \setlength\tabcolsep{1pt}
    {\DD
    \begin{tabular}{cccccccccc}
        GT &    GSA &   BT-H &   BDSD-PC &   PRACS &   MTF-GLP-FS &   MTF-GLP-HPM &   MTF-GLP-HPM-R &   AWLP &   MF \\
        \imagerr{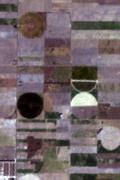} & \imagerr{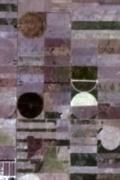} & \imagerr{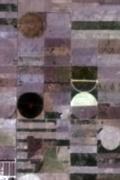} & \imagerr{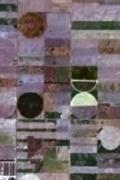} & \imagerr{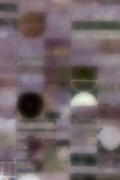} &
        \imagerr{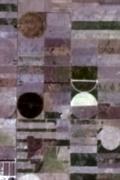} & \imagerr{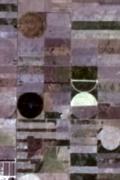} & \imagerr{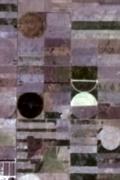} & 
         \imagerr{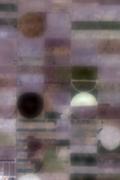} & \imagerr{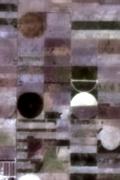} \\[1mm]
        HySURE &   SR-D &   TV &   HyperPNN &   HSpeNet &   DHP-DARN &   DIP-HyperKite &   Hyper-DSNet &   R-PNN &   PCA-Z-PNN \\
         \imagerr{20230908173127RR_HySure_RGB} & \imagerr{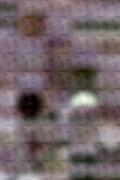} & \imagerr{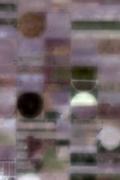} &
         \imagerr{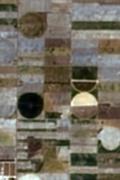} & \imagerr{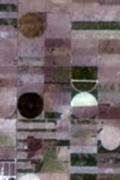} &
         \imagerr{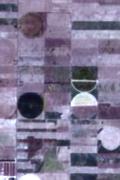} & \imagerr{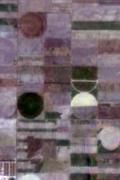} & \imagerr{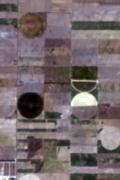} & \imagerr{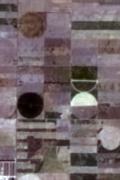} & \imagerr{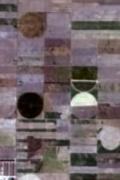} \\[1mm]
         \multicolumn{10}{c}{\D (a)} \\[1mm]
        GT &   GSA &   BT-H &   BDSD-PC &   PRACS &   MTF-GLP-FS &   MTF-GLP-HPM &   MTF-GLP-HPM-R &   AWLP &   MF \\
        \imagerr{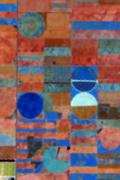} & \imagerr{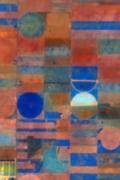} & \imagerr{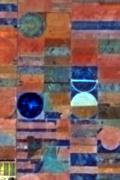} & \imagerr{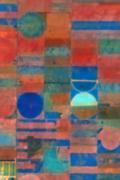} & \imagerr{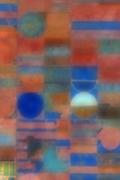} &
        \imagerr{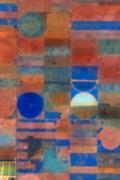} & \imagerr{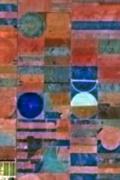} & \imagerr{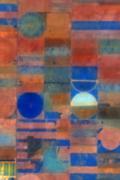} & 
         \imagerr{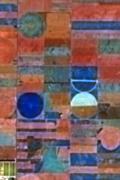} & \imagerr{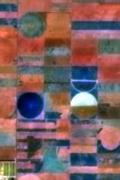} \\[1mm]
        HySURE &  SR-D &  TV &  HyperPNN &  HSpeNet &  DHP-DARN &  DIP-HyperKite &  Hyper-DSNet &  R-PNN &  PCA-Z-PNN \\
         \imagerr{20230908173127RR_HySure_FC} & \imagerr{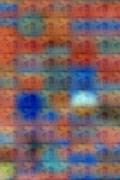} & \imagerr{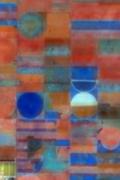} &
         \imagerr{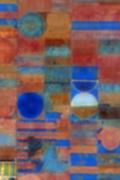} & \imagerr{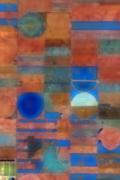} &
         \imagerr{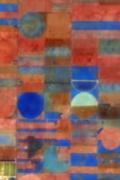} & \imagerr{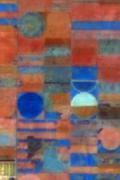} & \imagerr{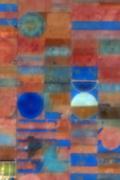} & \imagerr{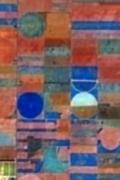} & \imagerr{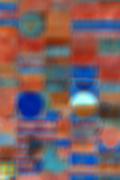} \\[1mm]
         \multicolumn{10}{c}{\D (b)} \\[1mm]
    \end{tabular}   
    }
    \caption{Pansharpening results (cropped: 120$\times$180) on Ford Country at reduced resolution. 
    Target GT (visible range on top, NIR-SWIR at bottom) 
    followed by all corresponding pansharpening results.
    Selected wavelenghts: $663, 560, 466$ nm (visible); $1943, 1261, 832$ nm (NIR-SWIR).}
    \label{fig:rr3}
\end{figure*}

\begin{figure*}
    \centering
    \setlength\tabcolsep{1pt}
    {\D
    \begin{tabular}{ccccccc}
        PAN &   HS &   GSA &   BT-H &   BDSD-PC &   PRACS &   MTF-GLP-FS \\
        \image{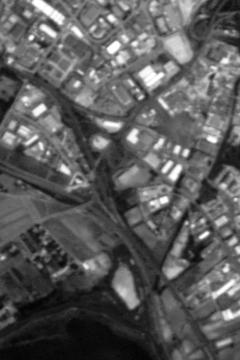} & \image{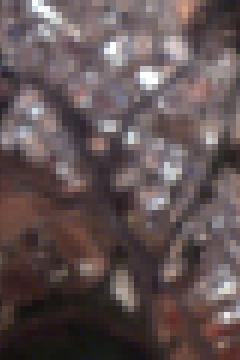} & \image{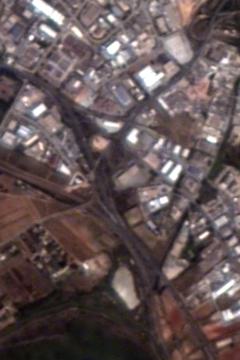} & \image{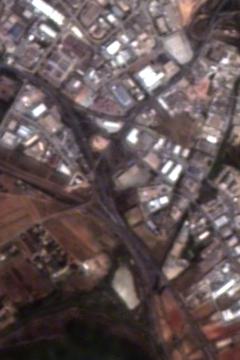} & \image{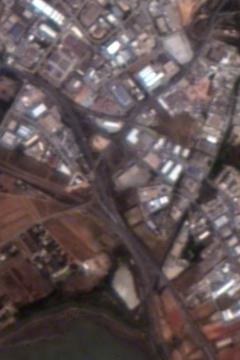} & \image{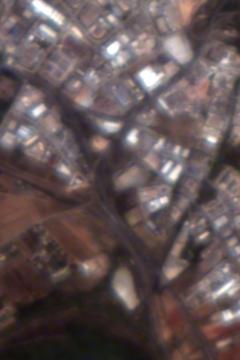} &\image{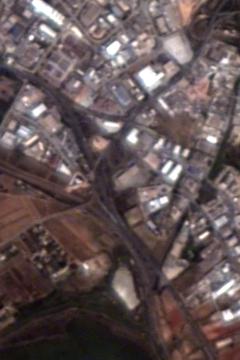} \\[1mm]
        MTF-GLP-HPM &   MTF-GLP-HPM-R &   AWLP &   MF &   HySURE &   SR-D &   TV\\
        \image{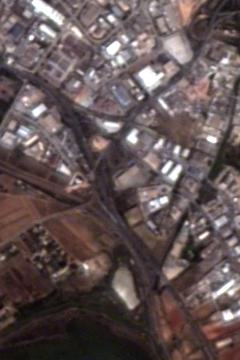} & \image{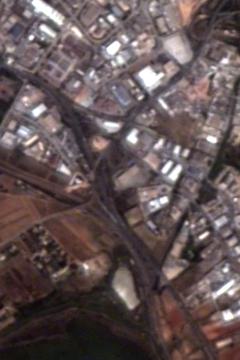} & 
         \image{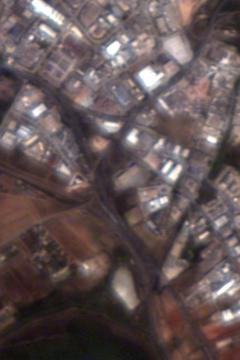} & \image{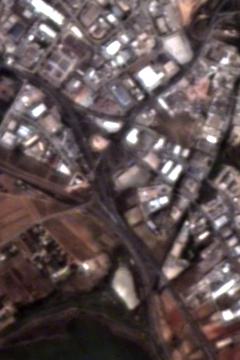} &
         \image{20220905101901FR_HySure_RGB} & \image{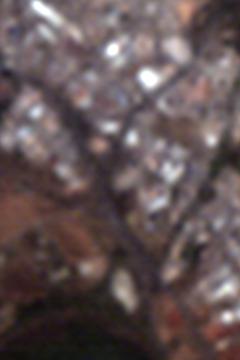} & \image{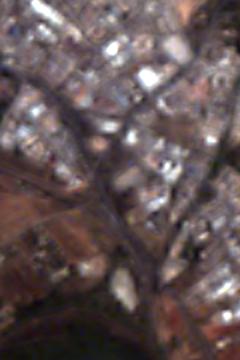}\\[1mm]
        HyperPNN &   HSpeNet &   DHP-DARN &   DIP-  HyperKite &   Hyper-DSNet &   R-PNN &   PCA-Z-PNN \\
         \image{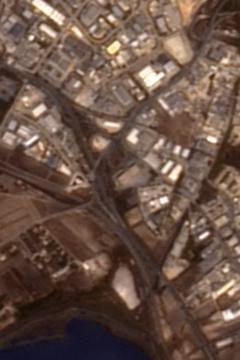} & \image{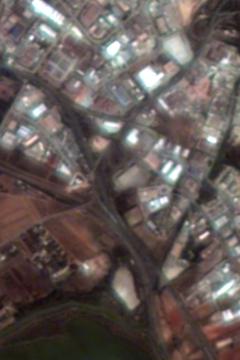} & \image{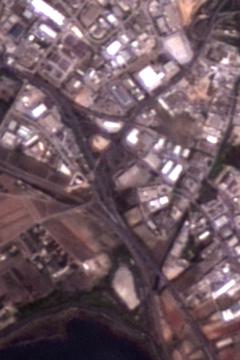} & \image{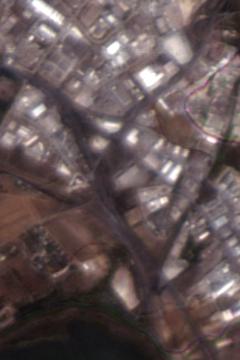} & \image{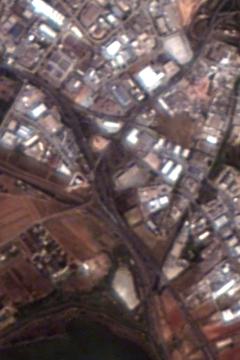} & \image{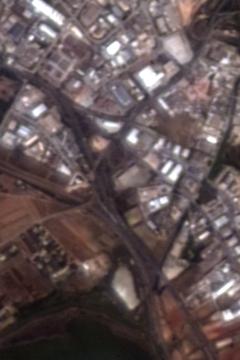} & \image{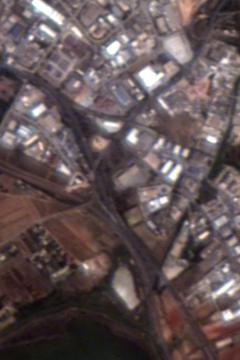} \\[1mm] 
    \end{tabular}   
    }
    \caption{Pansharpening results (cropped: 240$\times$360) on Cagliari at full resolution. 
    PAN image followed by the HS component 
    (bands in the visible range: $663, 560, 466$ nm; nearest-neighbor interpolation) 
    and all corresponding pansharpening results.}
    \label{fig:fr1-rgb}
\end{figure*}

\begin{figure*}
    \centering
    \setlength\tabcolsep{1pt}
    {\D
    \begin{tabular}{ccccccc}
        PAN &   HS &   GSA &   BT-H &   BDSD-PC &   PRACS &   MTF-GLP-FS \\
        \image{20220905101901FR_PAN_Z} & \image{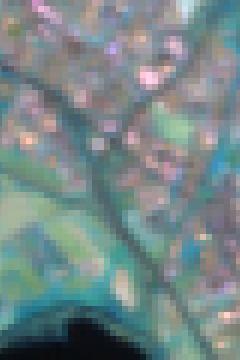} & \image{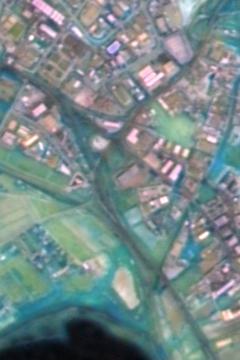} & \image{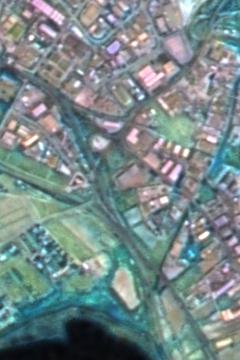} & \image{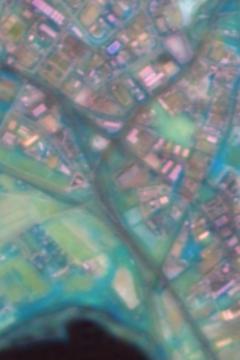} & \image{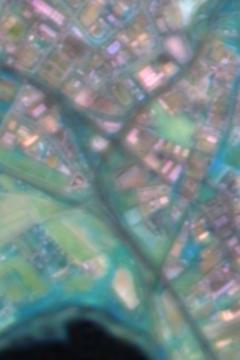} &\image{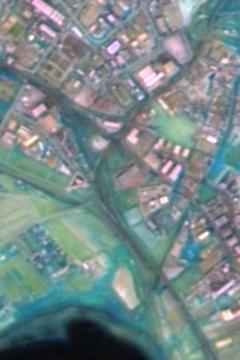} \\[1mm]
        MTF-GLP-HPM &   MTF-GLP-HPM-R &   AWLP &   MF &   HySURE &   SR-D &   TV\\
        \image{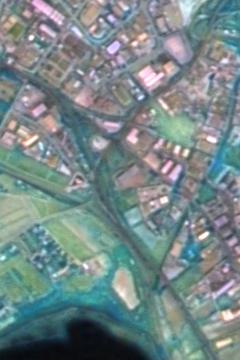} & \image{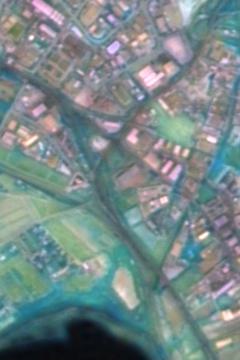} & 
         \image{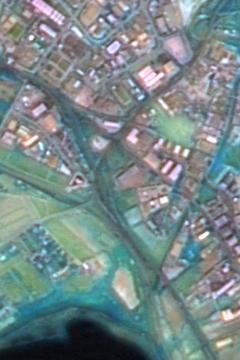} & \image{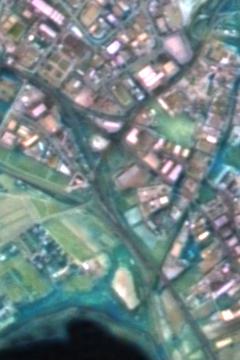} &
         \image{20220905101901FR_HySure_FC} & \image{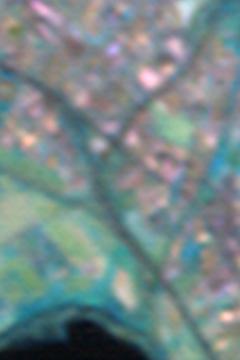} & \image{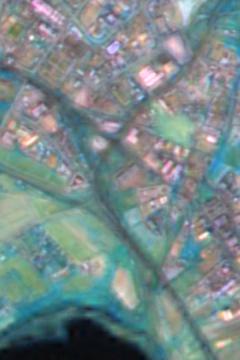}\\[1mm]
        HyperPNN &   HSpeNet &   DHP-DARN &   DIP-  HyperKite &   Hyper-DSNet &   R-PNN &   PCA-Z-PNN\\
         \image{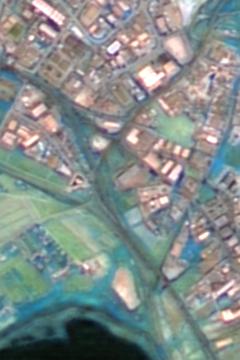} & \image{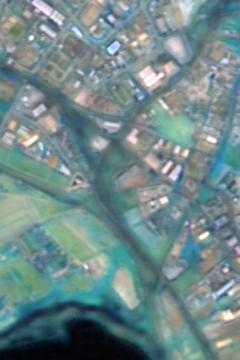} & \image{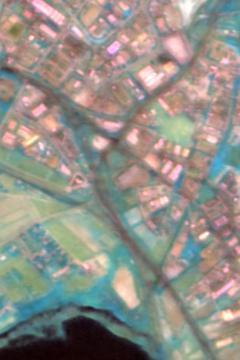} & \image{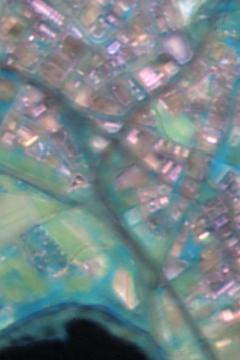} & \image{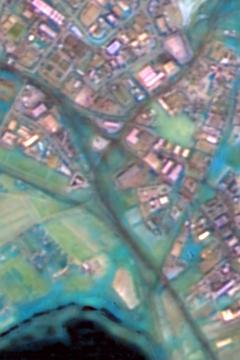} & \image{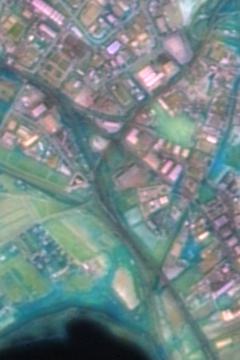} & \image{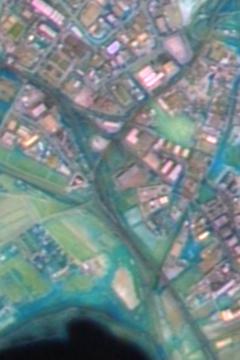} \\[1mm] 
    \end{tabular}
    }
    \caption{Pansharpening results (cropped: 240$\times$360) on Cagliari at full resolution. 
    PAN image followed by the HS component 
    (bands in the NIR-SWIR: $1943, 1261, 832$ nm; nearest-neighbor interpolation) 
    and all corresponding pansharpening results.}
    \label{fig:fr1-falsec}
\end{figure*}

\begin{figure*}
    
    \centering
    \setlength\tabcolsep{1pt}
    {\D
    \begin{tabular}{ccccccc}
        PAN &   HS &   GSA &   BT-H &   BDSD-PC &   PRACS &   MTF-GLP-FS \\
        \image{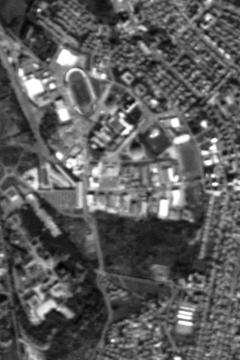} & \image{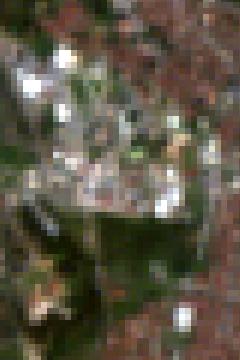} & \image{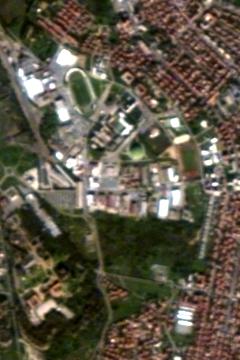} & \image{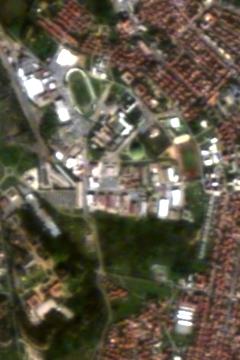} & \image{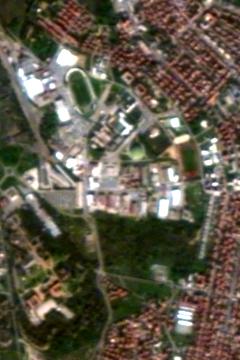} & \image{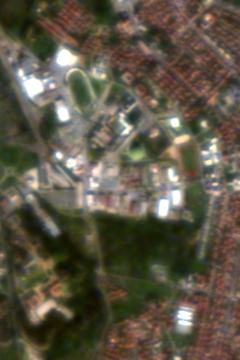} &\image{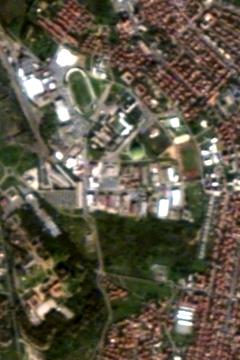} \\[1mm]
        MTF-GLP-HPM &   MTF-GLP-HPM-R &   AWLP &   MF &   HySURE &   SR-D &   TV\\
        \image{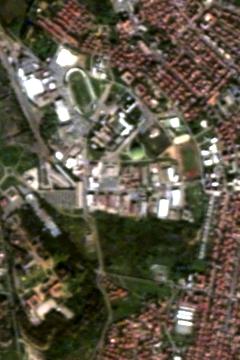} & \image{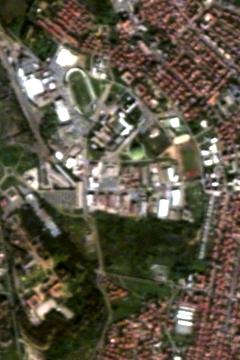} & 
         \image{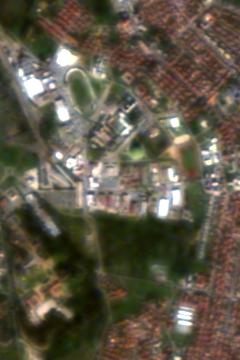} & \image{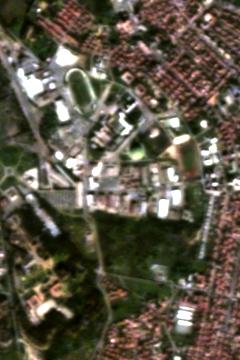} &
         \image{20231120102229FR_HySure_RGB} & \image{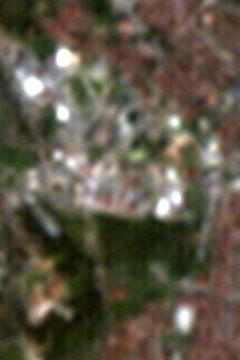} & \image{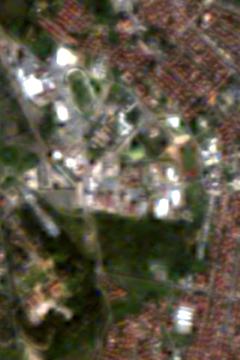}\\[1mm]
        HyperPNN &   HSpeNet &   DHP-DARN &   DIP-  HyperKite &   Hyper-DSNet &   R-PNN &   PCA-Z-PNN\\
         \image{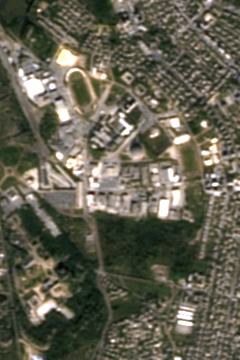} & \image{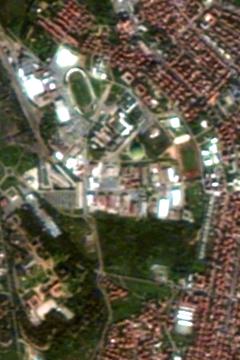} & \image{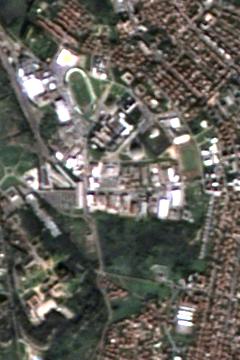} & \image{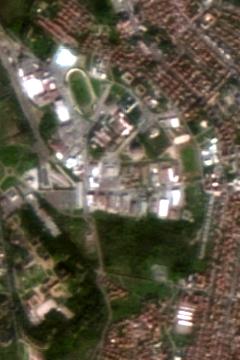} & \image{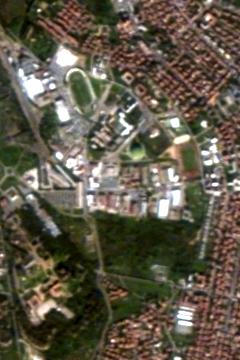} & \image{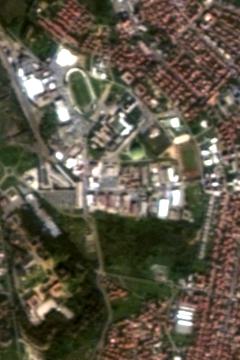} & \image{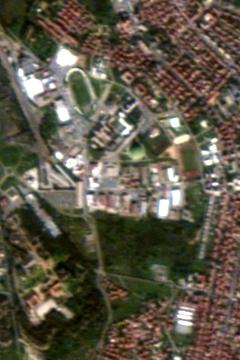} \\[1mm] 
    \end{tabular}    
    }
    \caption{Pansharpening results (cropped: 240$\times$360) on Macuspana (Tabasco) at full resolution. 
    PAN image followed by the HS component 
    (bands in the visible range: $663, 560, 466$ nm; nearest-neighbor interpolation) 
    and all corresponding pansharpening results.}
    \label{fig:fr4-rgb}
\end{figure*}

\begin{figure*}
    \centering
    \setlength\tabcolsep{1pt}
    {\D
    \begin{tabular}{ccccccc}
        PAN &   HS &   GSA &   BT-H &   BDSD-PC &   PRACS &   MTF-GLP-FS \\
        \image{20231120102229FR_PAN_Z} & \image{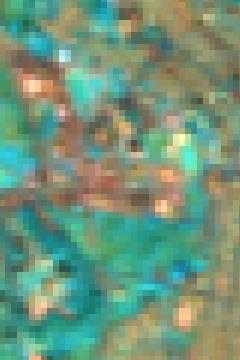} & \image{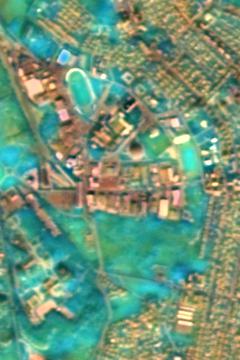} & \image{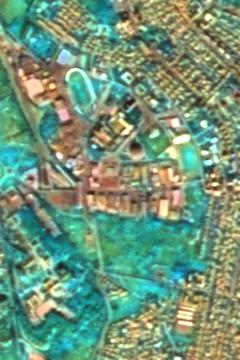} & \image{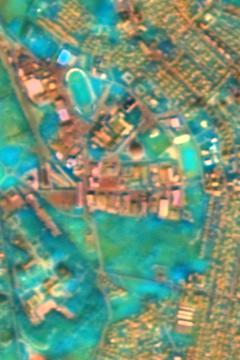} & \image{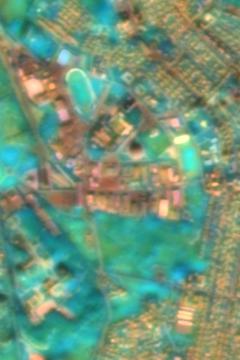} &\image{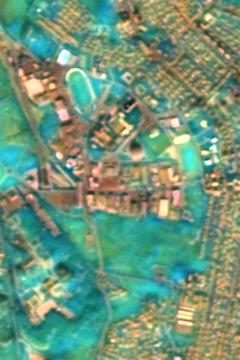} \\[1mm]
        MTF-GLP-HPM &   MTF-GLP-HPM-R &   AWLP &   MF &   HySURE &   SR-D &   TV\\
        \image{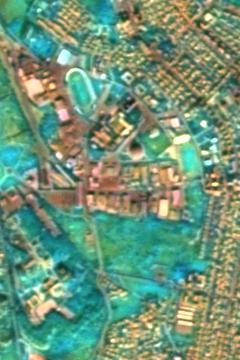} & \image{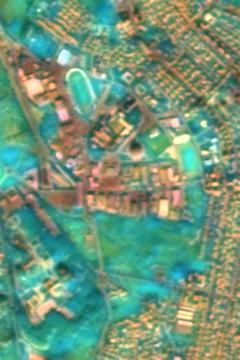} & 
         \image{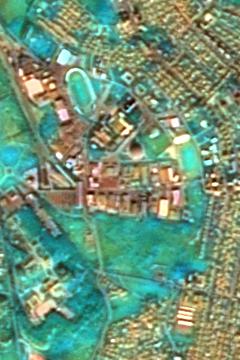} & \image{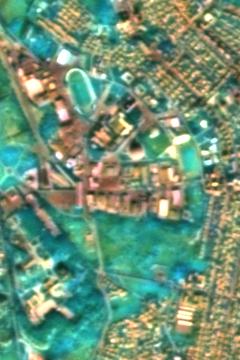} &
         \image{20231120102229FR_HySure_FC} & \image{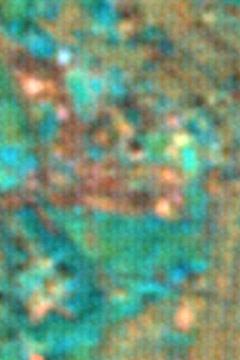} & \image{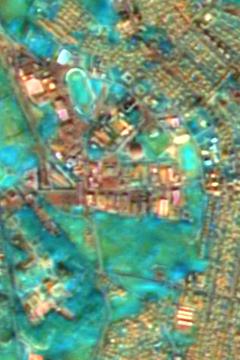}\\[1mm]
        HyperPNN &   HSpeNet &   DHP-DARN &   DIP-  HyperKite &   Hyper-DSNet &   R-PNN &   PCA-Z-PNN\\
         \image{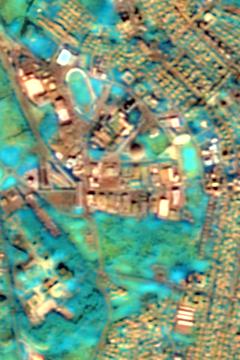} & \image{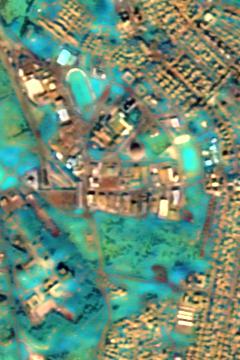} & \image{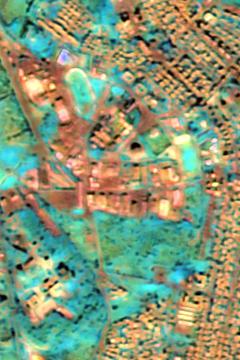} & \image{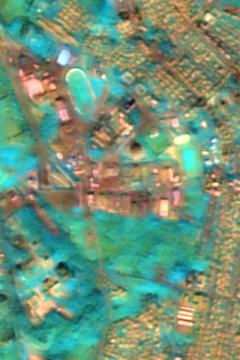} & \image{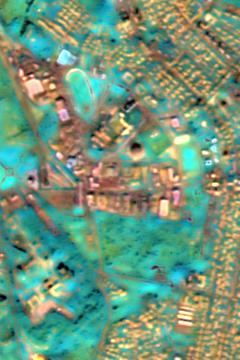} & \image{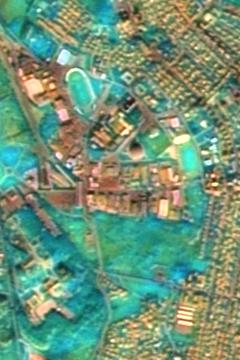} & \image{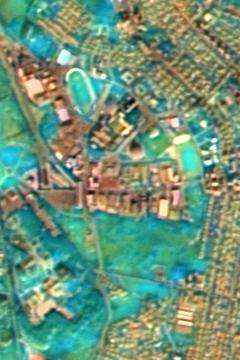} \\[1mm] 
    \end{tabular}    
    }
    \caption{Pansharpening results (cropped: 240$\times$360) on Macuspana (Tabasco) at full resolution. 
    PAN image followed by the HS component 
    (bands in the NIR-SWIR: $1943, 1261, 832$ nm; nearest-neighbor interpolation) 
    and all corresponding pansharpening results.
    }
    \label{fig:fr4-falsec}
\end{figure*}

\begin{figure}
    \centering
    \includegraphics{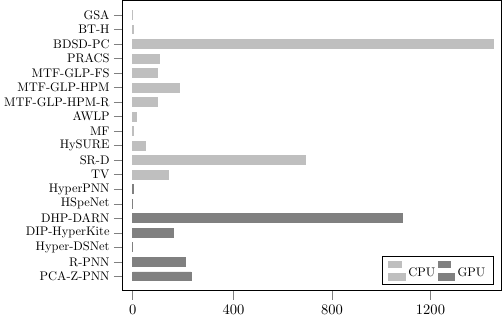}
    \caption{Average running time [seconds] per 600$\times$600 image (RR test images) at inference.
    DL methods are run on GPU, and the remaining ones on CPU.}
    \label{fig:time}
\end{figure}

\subsection{Generalization across datasets}
\label{sec:across_data}
To assess a method's ability to generalize across datasets, we fix the scale and index, {\em e.g.}, ERGAS, reduced resolution, and analyze score variations across images.
At reduced resolution (see Tab.~\ref{tab:RR}) virtuous examples are MTF-GLP-FS and MTF-GLP-HPM-R,
which exhibit high average scores for all three indexes, consistently across all test images.
The same also holds for some DL-based methods, {\em e.g.}, Hyper-DSNet, R-PNN, and others with worse average scores.
Examples of methods with generalization issues at reduced resolution are MTF-GLP-HPM or PCA-Z-PNN.
The former gets very good scores on Cagliari followed by bad ones on Udine.
The latter performs well on Tabasco, but much worse on Ford.

Moving to the full resolution results in Tab.~\ref{tab:FR},
consistent behaviour is observed across the four test scenes for almost all methods on the spectral distortion index $\DL$.
However, while all reduced resolution indices can be considered alternative measures of overall quality,
the full resolution indices $\DL$ and $\DS$ each focus on a specific feature, {\it i.e.} spectral or spatial quality.
Therefore, they are more likely to be uniform across images, following the characteristics of the method.
Inconsistencies are more easily observed in the RQNR hybrid index since it results from the combination of $\DL$ and $\DS$.
This is the case, for example, of BT-H or BDSD-PC, which show opposite spectral and spatial behaviours resulting in unstable RQNR rankings.

\subsection{Generalization across scales}
\label{sec:across_scales}

When moving from the reduced-resolution to the full-resolution domain,
image statistics change considerably, even if the resolution downgrade is carried out carefully accounting for the sensor MTF characteristics.
This is all the more true in our case, due to the rather large resolution ratio (six) of the PRISMA images.
Overall, the cross-scale statistical mismatch is more important than the cross-scene mismatch and cross-scale generalization\footnote{We note that
cross-scale invariance is somehow a proxy of cross-sensor invariance
because, even if the number/position of the spectral bands and the resolution ratio are the same, here the statistics change significantly, like for images acquired with different instruments.}
is not easily achieved.

To simplify the analysis of results, we consider only average scores,
focusing, for each method, on the coherence between rankings registered at RR with those occurring at FR.
The results of Tab.~\ref{tab:RR} and Tab.~\ref{tab:FR} fully confirm that the methods generalize much better across different scenes than across scales.
Some methods, like PRACS and PCA-Z-PNN, which struggle a bit at RR, seem to work well on FR data,
while others, in particular some supervised DL-based method, suffer a performance hit when switching to FR tests.
%Several other cases can be pointed out if the inspection is limited to SAM (for RR) and $\DL$ (for FR), the two spectral-oriented indexes.
Examples of methods that tend to perform fairly well (but not top) at both scales are GSA, Hyper-DSNet and R-PNN.
From these numbers, a picture emerges in which most methods seem to be optimized for just one scale, suffering on the other.
In some cases, these results are the effect of an explicit design choice, as in the case of supervised DL methods, where training must be carried out in the RR domain.
Given this apparent compromise, that is, working well only at one scale, one might wonder which one is preferable.
On the one hand, the ultimate goal is to work well on FR data.
On the other hand, at this scale, only consistency measures and not objective quality indicators can be observed.
Eventually, the question remains open.

%dead
%In fact, in the reduced resolution framework, ERGAS, SAM and $Q2^n$ do not tell apart the quality of low (coming from HS) and high (from PAN) spatial frequency components.
%The same SAM indicator which is more oriented to the spectral fidelity involves the whole range of spatial frequencies.
%As matter of fact (see Tab.~\ref{tab:RR}), the three RR indicators look quite correlated, with some exception, {\em e.g.} DHP-DARN
%which performs well on ERGAS and SAM, but not on $Q2^n$.

\subsection{Joint spectral-spatial quality}
\label{sec:spectral_spatial_balance}
In fusing the PAN and HS components we aim to keep both the spatial detail of the former and the spectral richness of the latter.
However, these goals are somewhat conflicting, and sometimes one property is obtained by sacrificing the other.
To analyze this trade-off we can refer to the FR consistency indices, $\DL$ and $\DS$, which evaluate spatial and spectral quality separately.
Indeed, the FR results of Tab.~\ref{tab:FR} highlight several situations where this undesirable behaviour emerges clearly,
with methods that show high scores on $\DL$ but poor performance on $\DS$, or {\it viceversa}.
The hybrid RQNR score, which weights the two indicators, can help evaluate the correct balance achieved by a method, with poor balance, signalled by a low RQNR, indicating ineffective fusion.
Experimental results show that some recent DL-based methods, like DIP-HyperKite, R-PNN and PCA-Z-PNN, offer excellent RQNR scores.
However, quite dated classical methods, like GSA and PRACS, are no less competitive.
This confirms the observation already made in the introduction, that DL has yet to realize its full potential for HS pansharpening.

All this said we must raise a warning on our quality metrics.
Spectral quality is relatively easy to assess, and Khan's index $\DL$ is quite robust and reliable.
On the contrary, the assessment of spatial quality remains an open problem as testified by several recent works on the topic \cite{Scarpa2022, Arienzo2022}.
Here, we decided to use $\DS$ in continuity with the prevalent literature and also with the recent HS pansharpening challenge \cite{Vivone2023}.
However, like other metrics, $\DS$ has well-known shortcomings and failure cases.
A good practice is to look first at the spectral quality, which can be reliably assessed through $\DL$, and then move on to spatial quality, 
complementing numerical indicators with the visual inspection of sample results, which might spot undetected anomalies and incoherencies.

\subsection{Subjective visual quality}
Since HS images comprise a very large number of spectral bands, examining them all individually would be unreasonable.
On the other hand, these bands form a limited number of groups with high interband correlation.
Therefore, as a reasonable compromise, we decided to show only six bands per sample image, 
grouped into two sets for false colour display, taken in the visible and NIR-SWIR ranges respectively.
The “visible” bands correspond roughly to the red, green and blue wavelengths.
The NIR-SWIR bands were selected such to have the minimum mutual correlation possible and be maximally representative of such a wide spectral range.
Furthermore, to save space and show images at a reasonable resolution for visual inspection, only a small but significant crop per image was selected for display (spotted with dashed boxes in Fig.~\ref{fig:prisma}).

For the RR framework we selected crops from Udine (see Fig.~\ref{fig:rr2}) and Ford Country (see Fig.~\ref{fig:rr3}), showing results for visible bands on the top and NIR-SWIR bands on the bottom.
On the top-left we show the ground truth (namely, the original HS), and then the pansharpening results for all methods.
For both scenes, visual inspection seems to confirm the indications provided by the numerical results of Tab.~\ref{tab:RR}.
Without going into too much detail, it can be easily verified that images that appear too blurred or present spectral distortion correspond to methods ranked in the worst positions in Tab.~\ref{tab:RR}.
On the other hand, given the availability of the GT, numerical assessment is very reliable in this domain.

In the FR framework, visual inspection turns out to be much more informative.
Here, lacking the GT, numerical results are less reliable as they only check consistency with the two input components, PAN and HS.
Visual results are displayed in Figures~\ref{fig:fr1-rgb}-\ref{fig:fr4-falsec} for crops taken from Cagliari and Tabasco, analyzed in the visible and NIR-SWIR ranges, respectively. 
In all cases, the original PAN and HS (the latter expanded to match sizes) are shown on the top left, followed by the output of all methods.
The comparison with the original HS shows some significant spectral deviations (colours do not match) like for HyperPNN and DHP-DARN on Cagliari.
In general, these problems were correctly predicted by the numerical indications of Tab.~\ref{tab:FR}).
In most cases, the spectral quality is definitely good, both in the visible and NIR-SWIR ranges, confirming that spectral fidelity is not the main issue in HS pansharpening.
The situation is more critical for spatial fidelity.
By comparing results with the PAN, several problems emerge, concerning blurring or, on the opposite side, oversharpening.
Some methods (PRACS, HySURE, SR-D, TV) produce systematic blurred outputs on both Cagliari and Tabasco and both in the visible and NIR-SWIR ranges.
In some other cases, this is a spectrally selective phenomenon:
for example, BDSD-PC blurs in the NIR-SWIR while AWLP blur in the visible range.
Finally, in a few cases (HSpeNet, DIP-HyperKite) the blur is image-dependent, occurring only on Cagliari.
In the opposite direction, some methods deliver over-sharpened output images, especially BT-H in the NIR-SWIR range.

These examples show that visual inspection plays a critical role in quality assessment, complementary to the role of numerical indicators.
Actually, the disagreement with numerical results can shed light on the reliability of performance metrics.
In particular, while $\DL$ seems to be consistent with the visual results,
$\DS$ shows some weakness as it does not always agree with the perceived spatial quality of the results.
This is the case, for example, of BDSD-PC and HySURE ($\DS\approx 0$) whose results look somewhat blurred both in visible and NIR-SWIR ranges and present failures on vegetated areas
as on the Tabasco image (Fig.~\ref{fig:fr4-rgb}).
Similar inconsistencies can also be observed for other solutions,
such as GSA and PRACS,
having moderately good $\DS$ values.

\subsection{Computational Time}

All experiments were carried out on the same server, an NVIDIA DGX Station A100, 
equipped with a 64-core AMD EPYC 7742 processor, 504 GB of DDR4 RAM, and four NVIDIA A100-SXM4 GPUs with 40 GB of GDDR5 memory each.
%featuring an AMD EPYC 7742 64-core Processor, 504 GB of DDR4 RAM, and four NVIDIA A100-SXM4 GPUs with 40GB of GDDR5 memory each.
The conventional CS, MRA, and MBO methods were run on the CPU, while the DL-based methods were run on a single GPU.
The average run time for each method on the 600$\times$600 RR test images is reported in Fig.~\ref{fig:time}.
The run times for the 4 times larger (1200$\times$1200) FR test images scale linearly with the size and hence are not reported.
However, computational scalability depends on the available hardware, so we do not draw general conclusions about that.
Some methods, both DL-based and not, have negligible run time, a quality to be considered in practical applications.
Others are exceedingly slow, such as the CPU-based BDSD-PC and the GPU-based DHP-DARN.
In general, all DL-based methods that try to adapt to the variability of input images pay a price in terms of computational complexity,
DHP-DARN and DIP-HyperKite to estimate the DIP, and R-PNN and PCA-Z-PNN for the target-adaptation phase.

\subsection{To what extent existing MS pansharpening methods generalize to the HS case?}

Given the nature of the problem at hand,
it is fair to ask to what extent a pansharpening method conceived for the PAN-MS fusion
can keep working satisfactory when replacing the MS with a HS datacube.
Skipping implementation issues due to the adaptation and computational
implications that differ from one method to another, 
a partial answer on quality can now be given on the basis of the achieved results.
The unique challenges present in the HS case are clear: 
much larger (and variable) number of bands;
spectral separation between the PAN and a large (the majority) number of bands;
larger resolution ratios to face ({\em e.g.}, 6 for PRISMA);
larger volume of data per geographic area.
%The problem of HS pansharpening, while appearing similar to MS one at first glance, presents unique challenges. Unlike MS and PAN images, which share a significant portion of the spectral range, HS data cover spectral frequencies well beyond those captured by the PAN sensor. Additionally, HS pansharpening typically involves different scale ratios and a considerably larger volume of data.

Despite these differences, the similarities between the two problems have led many researchers to apply MS pansharpening methods to HS data, either directly or with minor modifications. From the analysis of results, both quantitatively (refer to Table~\ref{tab:RR} and Table~\ref{tab:FR}) and qualitatively (refer to Figures~\ref{fig:fr1-rgb}-\ref{fig:fr4-falsec}), it appears that some MS pansharpening techniques show results comparable 
to those of specialized HS pansharpening solutions.
It is the case of methods such as MTF-GLP-FS, MTF-GLP-HPM-R, PRACS, GSA, TV. 
However, it is worth noticing that these methods, originally designed for a smaller number of bands that are well correlated with the PAN, 
may produce worse results on the bands beyond the PAN spectral bandwidth. 
For example, MTF-GLP-HPM performs well in the visible range but tends to produce oversharpened results in the NIR and SWIR wavelenghts. 
Similar observations can be made for methods such as BT-H and BDSD-PC.
The take home message is therefore that yes, MS-native solutions can be extended to the HS case taking care of the quality unbalance 
that may arise in the spectral domain.

\section{Conclusions}
\label{sec:conclusions}

In this work, we proposed a benchmarking framework for HS pansharpening, aiming to provide the research community with an easy-to-use tool to develop and test new methods at a faster pace.
Following a careful analysis of the SoTA, we have selected a series of reference methods, representative of the main and most promising approaches in the field.
All methods have been re-implemented and are ready for use.
In parallel, we designed a large dataset of real multiresolution PRISMA images (available on demand to the research community) to train and test all methods in a uniform environment and according to well-defined and stable protocols.
To establish a solid starting point for future research, we tested all selected methods using established performance metrics.
The analysis of the results allowed us to highlight the main problems of existing solutions and therefore the most promising research avenues.

In a way, we have taken a snapshot of the state of HS pansharpening research right now.
However, this is not intended to freeze things: on the contrary, it aims to encourage further research, allow easier comparison of results, and provide a context for sharing tools, as well as ideas.
In our vision, this framework is destined to expand with the help of interested researchers who can contribute new methods and data.

While claiming the merits of this initiative, we also see its limits.
The set of methods selected is certainly not exhaustive and all choices are to a certain extent arbitrary.
In particular, we could not reimplement some DL-based methods due to the lack of code and a sufficiently detailed description.
The dataset itself, while representing a notable improvement over existing ones, is limited to a single sensor, thus preventing true generalization across sensors from being tested.
Better performance parameters could probably be used to evaluate quality at full resolution.
In summary, there is much room for contributions and improvements.

\bibliographystyle{IEEEtran}
\bibliography{refs}
\end{document}